%% file: main.tex
\documentclass[10pt,twocolumn,letterpaper]{article}

\usepackage{cvpr}      

\input{preamble}

\usepackage{graphicx}
\usepackage{amsmath,amssymb} 
\usepackage{caption} 
\usepackage{wrapfig}
\usepackage{csquotes}
\usepackage{bm}
\usepackage{algorithm}
\usepackage[noend]{algpseudocode}
\usepackage{multirow}
\usepackage[misc]{ifsym}

\newcommand{\x}{{\bf x}}

\newcommand{\w}{{\bf w}}

\newcommand{\A}{\mathcal{A}}

\newcommand{\D}{\mathcal{D}}

\newcommand{\R}{\mathbb{R}}

\newcommand{\name}{{\sc Ease }}
\newcommand{\mame}{{\sc Ease}}
\newcommand{\ADAM}{{\sc Adam }}

\usepackage{accents}
\newlength{\dhatheight}

\usepackage{xcolor,colortbl}
\definecolor{Gray}{gray}{0.85}
\definecolor{Redo}{rgb}{0.95,0.69,0.51}
\definecolor{LightCyan}{rgb}{0.88,1,1}
\usepackage{afterpage}
\usepackage{pifont}

\usepackage{lipsum}
\definecolor{cvprblue}{rgb}{0.21,0.49,0.74}
\usepackage[pagebackref,breaklinks,colorlinks,citecolor=cvprblue]{hyperref}

\def\paperID{9306} 
\def\confName{CVPR}
\def\confYear{2024}

\title{Expandable Subspace Ensemble for \\Pre-Trained Model-Based Class-Incremental Learning  }

\author{Da-Wei Zhou, Hai-Long Sun, Han-Jia Ye\textsuperscript{(\Letter)}, De-Chuan Zhan\\
National Key Laboratory for Novel Software Technology, Nanjing University, China\\
School of Artificial Intelligence, Nanjing University, China\\
{\tt\small \{zhoudw,sunhl,yehj,zhandc\}@lamda.nju.edu.cn}
}

\begin{document}
\maketitle

\footnotetext[2]{Correspondence to: Han-Jia Ye (yehj@lamda.nju.edu.cn)}

\input{abstract}    
\input{intro}

\input{related}

\input{prelim}

\input{method}

\input{experiments}
\input{conclusion}

{
    \small
    \bibliographystyle{ieeenat_fullname}
    \bibliography{paper}
}

\setcounter{section}{0}
\renewcommand{\thesection}{\Roman{section}}
\begin{center}
	\textbf{\large Supplementary Material }
\end{center}
\setcounter{equation}{0}
\setcounter{figure}{0}
\setcounter{table}{0}
\setcounter{page}{1}
\makeatletter

\input{supp_material}

\end{document}

%% file: preamble.tex
\usepackage[dvipsnames]{xcolor}


%% file: abstract.tex
\begin{abstract}
	
Class-Incremental Learning (CIL) requires a learning system to continually learn new classes without forgetting. Despite the strong performance of Pre-Trained Models (PTMs) in CIL, a critical issue persists: learning new classes often results in the overwriting of old ones. Excessive modification of the network causes forgetting, while minimal adjustments lead to an inadequate fit for new classes. As a result, it is desired to figure out a way of efficient model updating without harming former knowledge.
In this paper, we propose ExpAndable Subspace Ensemble (\mame)
for PTM-based CIL. To enable model updating without conflict, we train a distinct lightweight adapter module for each new task, aiming to create task-specific subspaces. 
These adapters span a high-dimensional feature space, enabling joint decision-making across multiple subspaces. 
As data evolves, the expanding subspaces render the old class classifiers incompatible with new-stage spaces.
Correspondingly, we design a semantic-guided prototype complement strategy that synthesizes old classes' new features without using any old class instance. Extensive experiments on seven benchmark datasets verify \mame's state-of-the-art performance.
Code is available at: \url{https://github.com/sun-hailong/CVPR24-Ease}
\end{abstract}

%% file: intro.tex
\section{Introduction}

\begin{figure}[t]
	\vspace{-6mm}
	\begin{center}
		{\includegraphics[width=.95\columnwidth]{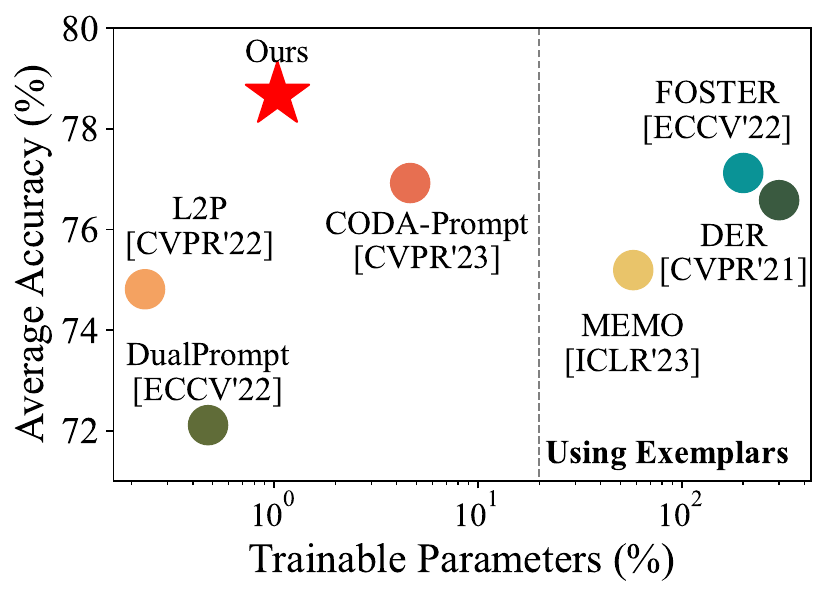}}
	\end{center}
	\vspace{-6mm}
	\caption{\small  Parameter-performance comparison of different methods on ImageNet-R B100 Inc50. All methods utilize the same PTM as initialization.
	\name requires the same scale parameters as other prompt-based methods~\cite{wang2022dualprompt,smith2023coda,wang2022learning} while performing best among all competitors without using exemplars.
	}
	\vspace{-7mm}
	\label{figure:parameter-size}
\end{figure}

The advent of deep learning leads to the remarkable performance of deep neural networks in real-world applications~\cite{deng2009imagenet,chen2022learning,chen2021large,ning2023rf,ye2019learning}. While in the open world, data often come in the stream format, requiring a learning system to incrementally absorb new class knowledge,  denoted as Class-Incremental Learning (CIL)~\cite{rebuffi2017icarl}.
 CIL faces a major hurdle: learning new classes tends to overwrite previously acquired knowledge, leading to catastrophic forgetting of existing features~\cite{french1999catastrophic,french1999modeling}.
Correspondingly, recent advances in pre-training~\cite{han2021pre} inspire the community to utilize pre-trained models (PTMs) to alleviate forgetting~\cite{wang2022learning,wang2022dualprompt}.
PTMs, pre-trained with vast datasets and substantial resources, inherently produce generalizable features.
Consequently, PTM-based CIL has shown superior performance, opening avenues for practical applications~\cite{wang2022s,smith2023coda,pian2023audio,villa2023pivot}.

With a generalizable PTM as initialization, algorithms tend to freeze the pre-trained weight and append minimal additional parameters (\eg, prompts~\cite{jia2022visual}) to accommodate incremental tasks~\cite{wang2022dualprompt,wang2022learning,wang2022s,smith2023coda}. Since pre-trained weights are frozen, the network's generalizability will be preserved along the learning process. 
Nevertheless, to capture new tasks' features, selecting and optimizing instance-specific prompts from the prompt pool inevitably rewrites prompts of former tasks.
Hence, it results in the conflict between old and new tasks,  triggering catastrophic  forgetting~\cite{jung2023generating}.

In CIL, the conflict between learning new knowledge and retaining old information is known as the {\em stability-plasticity dilemma}~\cite{grossberg2012studies}. 
Hence, learning new classes should not disrupt existing ones.
Several non-PTM-based methods, \ie, expandable networks~\cite{yan2021dynamically,wang2022foster,douillard2022dytox,chen2023dynamic}, address this by learning a distinct backbone for each new task, thereby creating a task-specific subspace. 
It ensures that optimizing a new backbone does not impact other tasks, and when concatenated, these backbones facilitate comprehensive decision-making across a high-dimensional space incorporating all task-specific features.
To map the concatenated features to corresponding classes,  a large classifier is optimized using {\em exemplars} \ie, instances of former classes.

Expandable networks resist the cross-task feature conflict, while they demand high resource allocation for backbone storage and necessitate the use of exemplars for unified classifier learning. In contrast, prompt learning enables CIL without exemplars but struggles with the forgetting of former prompts. This motivates us to question if it is possible to {\em construct low-cost task-specific subspaces to overcome cross-task conflict without the reliance on exemplars}. 

There are two main challenges to achieving this goal. {\bf 1)} Constructing low-cost, task-specific subspaces.  Since tuning PTMs requires countless resources, we need to create and save task-specific subspaces with lightweight modules instead of the entire backbone. 
{\bf 2)} Developing a classifier that can map continuously expanding features to corresponding classes.
Since exemplars from former stages are unavailable, the former stages' classifiers are incompatible with continual-expanding features. 
Hence, we need to utilize the class-wise relationship as semantic guidance to {\em synthesize} the classifiers of formerly learned classes.

In this paper, we propose ExpAndable Subspace Ensemble (\mame) to tackle the above challenges. 
To alleviate cross-task conflict, we learn task-specific subspace for each incremental task, making learning new classes not harm former ones. 
These subspaces are learned by adding lightweight adapters based on the frozen PTM, so the training and memory costs are negligible.
Hence, we can concatenate the features of PTM with every adapter to aggregate information from multiple subspaces for a holistic decision. 
Moreover, to compensate for the dimensional mismatch between existing classifiers and expanding features, we utilize class-wise similarities in the co-occurrence space to guide the classifier mapping in the target space. Thus, we can synthesize classifiers of former stages without using exemplars. 
During inference, we reweight the prediction result via the compatibility between features and prototypes and build a robust ensemble considering the alignment of all subspaces.
As shown in Figure~\ref{figure:parameter-size}, \name shows state-of-the-art performance with limited memory cost.

%% file: related.tex
\section{Related Work}
 
\noindent\textbf{Class-Incremental Learning (CIL)}: requires a learning system to continually absorb new class knowledge without forgetting existing ones~\cite{wang2024few,zhuang2023gkeal,zhuang2022acil,liu2021rmm,zhao2021mgsvf,dong2022federated,dong2023federated,gao2022rdfcil,wang2023beef,goswami2024fecam}, which can be roughly divided into several categories. Data rehearsal-based methods~\cite{aljundi2019gradient,liu2020mnemonics,ratcliff1990connectionist,zhao2021memory,chaudhry2018efficient} select and replay exemplars from former classes when learning new ones to recover former knowledge. Knowledge distillation-based methods~\cite{li2017learning,rebuffi2017icarl,douillard2020podnet,zhang2020class,simon2021learning,tao2020topology,dhar2019learning} build the mapping between the former stage model and the current model via knowledge distillation~\cite{hinton2015distilling}. The mapped logits/features help the incremental model to reflect former characteristics during updating. Parameter regularization-based methods~\cite{kirkpatrick2017overcoming,aljundi2019task,aljundi2018memory,zenke2017continual} exert regularization terms on the drift of important parameters during model updating to maintain former knowledge. Model rectification-based methods~\cite{zhao2020maintaining,wu2019large,yu2020semantic,shi2022mimicking,belouadah2019il2m,pham2021continual} correct the inductive bias of incremental models for unbiased prediction. Recently, expandable networks~\cite{yan2021dynamically,wang2022foster,douillard2022dytox,chen2023dynamic,hu2023dense,huang2023resolving} show strong performance among other competitors. Facing a new incremental task, they keep the previous backbone in the memory and initialize a new backbone to capture these new features. As for prediction, they concatenate all the backbones for a large feature map and learn a corresponding classifier with extra exemplars to calibrate among all classes. There are two main reasons that hinder the deployment of model expansion-based methods in pre-trained model-based CIL, \ie, the huge memory cost for large pre-trained models and the requirement of exemplars.

\noindent\textbf{Pre-Trained Model-Based CIL}: is now a hot topic in today's CIL field~\cite{zhou2024continual,wang2024hierarchical,mcdonnell2024premonition}. With the prosperity of pre-training techniques, it is intuitive to introduce PTMs into CIL for better performance. Correspondingly, most methods~\cite{wang2022dualprompt,wang2022learning,smith2023coda,wang2022s} learn a prompt pool to adaptively select the instance-specific prompt~\cite{jia2022visual} for model updating. With the pre-trained weights frozen, these methods can encode new features into the prompt pool.
DAP~\cite{jung2023generating} further extends the prompt selection process with a prompt generation module. Apart from prompt tuning, LAE~\cite{gao2023unified} proposes EMA-based model updating with online and offline models. SLCA~\cite{zhang2023slca} extends the Gaussian modeling of previous classes in~\cite{zhu2021prototype} to rectify classifiers during model updating. Furthermore, ADAM~\cite{zhou2023revisiting} shows that prototypical classifier~\cite{snell2017prototypical} is a strong baseline, and RanPAC~\cite{mcdonnell2023ranpac} explores the application of random projection in this setting.

%% file: prelim.tex
\section{Preliminaries}
In this section, we introduce the background of class-incremental learning and pre-trained model, baselines, and their limitations.

\subsection{Class-Incremental Learning}
CIL is the learning scenario where a model continually learns to classify new classes to build a unified classifier~\cite{rebuffi2017icarl}. 
Given a sequence of $B$ training sets, denoted as $\left\{\D^{1}, \D^{2}, \cdots, \D^{B}\right\}$, where $\D^{b}=\left\{\left(\x_{i}, y_{i}\right)\right\}_{i=1}^{n_b}$ is the $b$-th training set with $n_b$ instances.
An instance $\x_i \in \R^D$ is from class $y_i \in Y_b$. $Y_b$ is the label space of task $b$, and $Y_b  \cap Y_{b^\prime} = \varnothing$ for $b\neq b^\prime$, \ie, non-overlapping classes for different tasks.
We follow the {\bf exemplar-free setting} in~\cite{wang2022learning,wang2022dualprompt,smith2023coda}, where we save no exemplars from old classes.
Hence, during the $b$-th incremental stage,  we can only access data from $\D^b$ for training.
In CIL, we aim to build a unified classifier for all seen classes $\mathcal{Y}_b=Y_1 \cup \cdots Y_b$ as data evolves. Specifically, we hope to find a model $f(\x): X\rightarrow\mathcal{Y}_b$ that minimizes the expected risk:
\begin{equation} \label{eq:totalrisk} 
	f^*=\underset{f \in \mathcal{H}}{\operatorname{argmin}} \; \mathbb{E}_{(\mathbf{x}, y) \sim \mathcal{D}_{t}^1\cup\cdots\mathcal{D}_{t}^b} \mathbb{I}\left(y \neq f(\mathbf{x})\right) \,.
\end{equation}
$\mathcal{H}$ is the hypothesis space and $\mathbb{I}(\cdot)$ denotes the indicator function. $\mathcal{D}_{t}^b$ represents the data distribution of task $b$. Following typical PTM-based CIL works~\cite{wang2022learning,wang2022dualprompt,smith2023coda}, we assume that a pre-trained model (\eg, Vision Transformer~\cite{dosovitskiy2020image}) is available as the initialization for $f(\x)$. We decouple the PTM into the feature embedding $\phi(\cdot):\R^D\rightarrow \R^d$ and a linear classifier $W\in \R^{d\times|\mathcal{Y}_{b}|}$. The embedding function $\phi(\cdot)$ refers to the final $\texttt{[CLS]}$ token in ViT, and the model output is denoted as $f(\x)=W^\top\phi(\x)$. For clarity, we decouple the classifier into $W=[\w_1,\w_2,\cdots,\w_{|\mathcal{Y}_{b}|}]$, and the classifier weight for class $j$ is $\w_j$.

\subsection{Baselines in Class-Incremental Learning} \label{sec:baseline}

\noindent\textbf{Learning with PTMs:} In the era of PTMs, many works~\cite{wang2022dualprompt,wang2022learning,smith2023coda,wang2022s,jung2023generating} seek to modify the PTM {\em slightly}, in order to maintain the pre-trained knowledge. The general idea is to freeze the pre-trained weights and train the learnable prompt pool (denoted as $\mathbf{Pool}$) to influence the self-attention process and encode task information. Prompts are learnable tokens with the same dimension as image patch embedding~\cite{dosovitskiy2020image,jia2022visual}. The  target is formulated as:
\begin{equation} \label{eq:l2p} 
	\min_{\mathbf{Pool}\cup W}\sum_{(\x,y)\in{D^b}} \ell \left(W^\top
	\bar\phi\left(\x; \mathbf{Pool} \right)	, y\right)+ \mathcal{L}_{\mathbf{Pool}} \,,
\end{equation}
where $\ell(\cdot,\cdot)$ is the cross-entropy loss that measures the discrepancy between prediction and ground truth.
$\mathcal{L}_{\mathbf{Pool}}$ denotes the prompt selection~\cite{wang2022learning} or regularization~\cite{smith2023coda} term for prompt training. 
Optimizing Eq.~\ref{eq:l2p} encodes the task information into these prompts, enabling the PTM to capture more class-specific information as data evolves. 

\noindent\textbf{Learning with expandable backbones:}  Eq.~\ref{eq:l2p} enables the continual learning of a pre-trained model, while training prompts for new classes will conflict with old ones and lead to forgetting.
 Before introducing PTMs to CIL, methods consider model expansion~\cite{yan2021dynamically,wang2022foster} to tackle cross-task conflict. Specifically, when facing an incoming task, the model freezes the previous backbone $\bar{\phi}_{old}$ and keeps it in memory, and initializes a new backbone $\phi_{new}$. Then it aggregates the embedding functions $[\bar{\phi}_{old}(\cdot), \phi_{new}(\cdot)]$ and initializes a larger fully-connected layer $W_{E}\in\R^{2d\times|\mathcal{Y}_{b}|}$. During updating, it optimizes the cross-entropy loss to train the new embedding and classifier:
\begin{equation} \label{eq:der}
	\min_{\phi_{new}\cup W_E} \sum_{(\x,y)\in{D^b\cup \mathcal{E}}}\ell (W_{E}^\top [\bar{\phi}_{old}(\x),\phi_{new}(\x)],y) \,,
\end{equation}
where $\mathcal{E}$ is the {\bf exemplar} set containing instances of former classes (which is {\em unavailable} in the current setting). Eq.~\ref{eq:der} depicts a way to learn new features for new classes. Assuming the first task contains `cats,' the old embedding will be tailored for extracting features like beards and stripes due to limited model capacity. If the incoming task contains `birds,' instead of erasing the former features in $\phi_{old}$, Eq.~\ref{eq:der} resorts to a new backbone $\phi_{new}$ to capture features like beaks and feathers. The concatenated features enable the model to learn new features without harming old ones, and the model calibrates among all seen classes by tuning a classifier with the exemplar set.

\noindent\textbf{Learning expandable subspaces for PTM:} 
  Eq.~\ref{eq:l2p} encodes the task information into the prompts while optimizing prompts for new tasks will result in conflict with old ones. 
 By contrast, expanding backbones reveal a promising way to alleviate cross-task overwriting while the model scale and computational cost of PTMs hinder the application of Eq.~\ref{eq:der} in PTM-based CIL.
 Additionally, since we do not have any exemplars $\mathcal{E}$, optimizing Eq.~\ref{eq:der} also fails to achieve a well-calibrated classifier for all seen classes.
Hence, this inspires us to explore whether {\em it is possible to achieve low-cost subspace expansion without using exemplars}.

%% file: method.tex
\begin{figure*}[t]
		\vspace{-7mm}
	\begin{center}
		{\includegraphics[width=1.98\columnwidth]{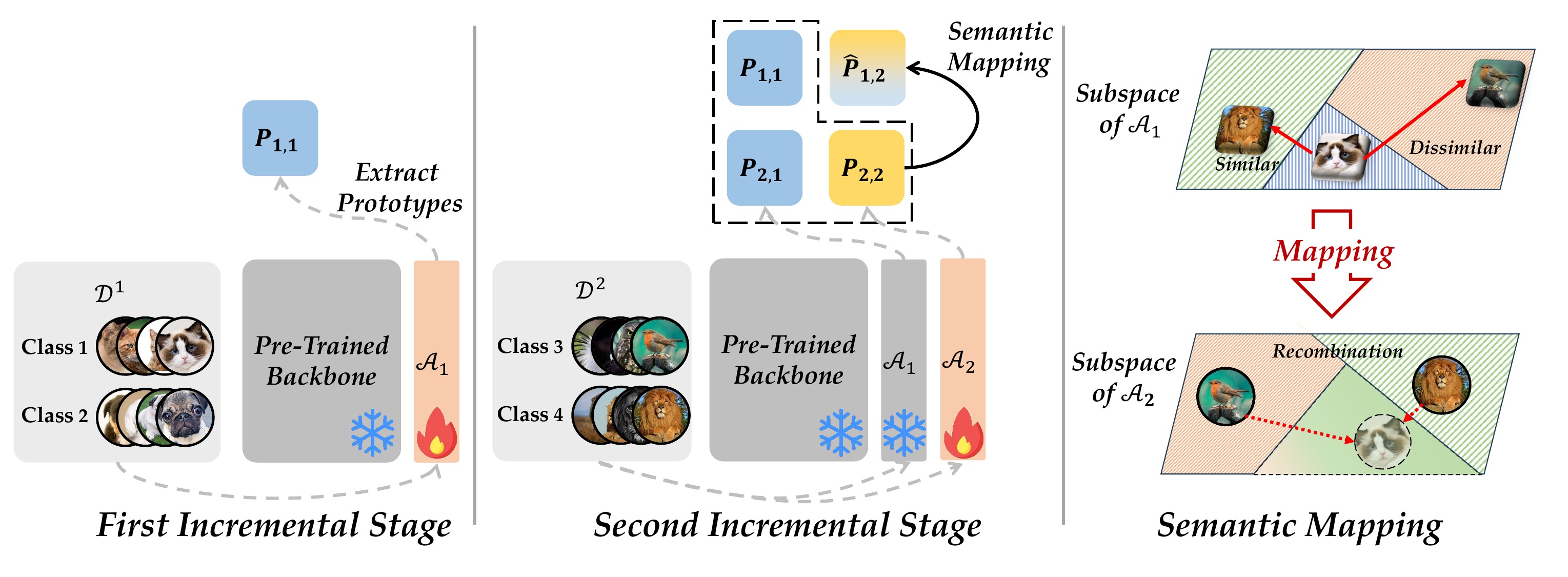}}
	\end{center}
		\vspace{-7mm}
	\caption{\small  Illustration of \mame. {\bf Left}: In the first task, we learn an adapter $\A_1$ to encode task specific features, and extract class prototypes $\mathbf{P}_{1,1}$. {\bf Middle}: In the second task, we initialize a new adapter $\A_2$ to encode new features, and extract prototypes $\mathbf{P}_{2,1}$ and $\mathbf{P}_{2,2}$. Without exemplars, we need to synthesize $\mathbf{P}_{1,2}$ (old class prototypes in the new subspace) for prediction.  {\bf Right}: Semantic mapping process. We extract class-wise similarity in the co-occurrence subspace and utilize it to synthesize old class prototypes in the target space.
	}
		\vspace{-5mm}
	\label{figure:teaser}
\end{figure*}

\section{{\scshape{Ease}}: Expandable Subspace Ensemble}

Observing that subspace expansion can potentially mitigate cross-task conflict in CIL, we aim to achieve this goal {\em without exemplars}. Hence, we first create lightweight subspaces for sequential tasks to control the total budget and computational cost. The adaptation modules should reflect the task information to provide task-specific features so that learning new tasks will not harm former knowledge. 
On the other hand, since we do not have exemplars, we are unable to train a classifier for the ever-expanding features. Hence,  we need to {\em synthesize and complete} the expanding classifier and calibrate the predictions among different tasks without using historical instances. Correspondingly, we attempt to utilize semantic-guided mapping to complete former classes in the latter subspace.  Afterward, the model can enjoy the strong generalization ability of the pre-trained model and various task-specific features in a unified high-dimensional decision space and make the predictions holistically without forgetting existing ones.

We first introduce the subspace expansion process and then discuss how to complete the classifiers. We summarize the inference function with pseudo-code in the last part.

\subsection{Subspace Expansion with Adapters}

In Eq.~\ref{eq:der}, new embedding functions are obtained through fully finetuning the previous model. However, it requires a large computational cost and memory budget to finetune and save all these backbones. By contrast, we suggest achieving this goal through lightweight adapter tuning~\cite{houlsby2019parameter,chenadaptformer}. Denote there are $L$ transformer blocks in the pre-trained model, each containing a self-attention module and an MLP layer. Following~\cite{chenadaptformer}, we learn an adapter module as a side branch for the MLP. Specifically, an adapter is a bottleneck module that contains a down-projection layer $W_{down}\in\R^{d\times r}$, a non-linear activation function $\sigma$, and an up-projection layer $W_{up}\in\R^{r\times d}$. It adjusts the output of the MLP as:
\begin{equation} \label{eq:adapter_format}
	\x_o=\sigma(\x_iW_{down})W_{up}+\text{MLP}(\x_i)\,,
\end{equation}
where $\x_i$ and $\x_o$ represent the input and output of MLP, respectively.
Eq.~\ref{eq:adapter_format} reflects the task information by adding the residual term to the original output.
 We denote the set of adapters among all $L$ transformer blocks as $\mathcal{A}$ and the adapted embedding function with adapter $\mathcal{A}$ as $\phi(\x;\A)$. Hence, facing a new incremental task, we can freeze the pre-trained weights and only optimize the adapter by:
\begin{equation}  \label{eq:adapter}
  \min_{\A\cup W} \sum_{(\x,y)\in \D^b} \ell \left(W^\top
  \bar\phi\left(\x; \mathcal{A} \right)	, y\right) \,.
\end{equation}
Optimizing Eq.~\ref{eq:adapter} enables us to encode task-specific information in these lightweight adapters and create task-specific subspaces. Correspondingly, we share the frozen pre-trained backbone and learn {\em expandable adapters for each new task}. During the learning process of task $b$, we initialize a new adapter $\A_b$ and optimize Eq.~\ref{eq:adapter} to learn task-specific subspaces. This results in a list of $b$ adapters: $\{\A_1,\A_2,\cdots,\A_b\}$. Hence, we can easily get the concatenated features in all subspaces by concatenating the pre-trained backbone with every adapter:
\begin{equation} \label{eq:expand_adapter}
	\Phi(\x)=[\phi(\x;\A_1), \cdots, \phi(\x;\A_b)]\in \R^{bd} \,.
\end{equation}
\noindent\textbf{Effect of expandable adapters}: Figure~\ref{figure:teaser} (left and middle) illustrates the adapter expansion process. 
Since we only tune the task-specific adapter with the corresponding task, training the new task will not harm the old knowledge (\ie, former adapters).
Moreover, in Eq.~\ref{eq:expand_adapter}, we combine the pre-trained embedding with various task-specific adapters to get the final presentation. The embedding contains all task-specific information in various subspaces that can be further integrated for a holistic prediction. Furthermore, since adapters are only lightweight branches, they require much fewer parameters than fully finetuning the backbone. The parameter cost for saving these adapters is $(B\times L\times 2dr)$, where $B$ is the number of tasks, $L$ is the number of transformer blocks, and $2dr$ denotes the parameter number of each adapter (\ie, linear projections). 

After getting the holistic embedding, we discuss how to build the mapping from $b d$ dimensional features to classes. 
We utilize a prototype-based classifier~\cite{snell2017prototypical} for prediction. Specifically, after the training process of each incremental stage, we extract the class prototype of the $i$-th class in adapter $\A_b$'s subspace:
\begin{equation}  \label{eq:prototype}\textstyle
	{\bm p}_{i,b}= \frac{1}{N} \sum_{j=1}^{|\mathcal{{D}}^b|}\mathbb{I}(y_j=i)\phi(\x_j;\A_b) \,,
\end{equation} 
where $N$ is the instance number of class $i$. Eq.~\ref{eq:prototype} denotes the most representative pattern of the corresponding class in the corresponding embedding space, and we can utilize the concatenation of prototypes in all adapters' embedding spaces $\mathcal{P}_i=[{\bm p}_{i,1},{\bm p}_{i,2},\cdots,{\bm p}_{i,b}]\in \R^{b d}$ to serve as class $i$'s classifier.
Hence, the classification is based on the similarity of a corresponding embedding $\Phi(\x)$ and the concatenated prototype, \ie, $p(y|\x) \propto \text{sim}\langle \mathcal{P}_y,\Phi(\x)\rangle$. We utilize a cosine classifier for prediction. 

\subsection{Semantic Guided Prototype Complement}

Eq.~\ref{eq:prototype} builds classifiers with representative prototypes. 
However, when a new task arrives, we need to learn a new subspace with a new adapter. It requires recalculating all class prototypes in the {\em latest subspace} to align the prototypes with the increasing embeddings, while we do not have any exemplars to estimate that of old classes. 
For example, we train $\A_1$ with the first dataset $\D^1$ in the first stage and extract prototypes for classes in $\D^1$, denoted as $\mathbf{P}_{1,1}=\text{Concat}[{\bm p}_{1,1};\cdots {\bm p}_{|\mathcal{Y}_{1}|,1}]\in\R^{|\mathcal{Y}_{1}|\times d }$. The former subscript in $\mathbf{P}_{1,1}$ stands for the task index, and the latter for the subspace. 
 In the following task, we expand an adapter $\A_2$ with $\D^2$. Since we only have $\D^2$, we can only calculate prototypes of $\D^2$ in $\A_1$ and $\A_2$'s subspaces, \ie,  $\mathbf{P}_{2,1}$, $\mathbf{P}_{2,2}$. In other words, we cannot calculate the prototypes of old classes in the new embedding space, \ie, $\mathbf{P}_{1,2}$. 
 This results in the {\em inconsistent dimension} between prototypes and embeddings, and we need to find a way to complete and synthesize prototypes of old classes in the latest subspace.

 Without loss of generality, we formulate the above problem as: given two subspaces (old and new) and two class sets (old and new), the target is to estimate old class prototypes in the new subspace $\hat{\mathbf{P}}_{o,n}$ using $\mathbf{P}_{o,o}$, $\mathbf{P}_{n,o}$, $\mathbf{P}_{n,n}$.
 Among them, $\mathbf{P}_{o,o}$ and $\mathbf{P}_{n,o}$ represent prototypes of old and new classes in the old subspace (which we call co-occurrence space), and $\mathbf{P}_{n,n}$ represents new classes prototypes in the new subspace.

 Since related classes rely on similar features to determine the label, it is intuitive to reuse similar classes' prototypes to synthesize a prototype of a related class. For example, essential  features representing  a `lion' can also help define  a `cat.'
 We consider such semantic similarity can be shared among different embedding spaces, \ie, the similarity between `cat' and `lion' should be shared across different adapters' subspaces. 
 Hence, we can extract such {\em semantic information} in the co-occurrence space and restore the prototypes by recombining related prototypes. 
 Specifically, we measure the similarity between old and new classes in the old subspace (where all classes co-occur) and utilize it to reconstruct prototypes in the new embedding space. The class-wise similarity among classes is calculated via prototypes in the co-occurrence subspace:
 \begin{equation} \label{eq:sim}
 	\text{Sim}_{i,j}=\frac{\mathbf{P}_{o,o}[i]}{\|\mathbf{P}_{o,o}[i]\|_2} \frac{\mathbf{P}_{n,o}[j]^\top}{\|\mathbf{P}_{n,o}[j]\|_2} \,,
 \end{equation}
 where the index $i$ denotes the $i$-th class's prototype.
 In Eq.~\ref{eq:sim}, we measure the semantic similarity of an old class prototype to a new class prototype in the same subspace and get the similarity matrix. We further normalize the similarities via softmax: $
 	{\text{Sim}}_{i,j} = \frac{\exp^{\text{Sim}_{i,j}}}{\sum_j \exp^{\text{Sim}_{i,j}}}$. The normalized similarity denotes the local relative relationship of an old class to new classes in the co-occurrence space, which is supposed to be shared across different subspaces.
 	
 After getting the similarity matrix, we further utilize the relative similarity to reconstruct old class prototypes in the new subspace. Since the relationship between classes can be shared among different subspaces, the value of old class prototypes can be measured by the weighted combination of new class prototypes:
 \begin{equation} \label{eq:reconstruct} \textstyle
  \hat{\mathbf{P}}_{o,n}[i]=\sum_j  \text{Sim}_{i,j} \times \mathbf{P}_{n,n}[j] \,.
 \end{equation} 
 \noindent\textbf{Effect of prototype complement:} 
 Figure~\ref{figure:teaser} (right) depicts the prototype synthesis process.
 With Eq.~\ref{eq:reconstruct}, we can restore the old class prototypes in the latest subspace without any former exemplars.   After learning each new adapter, we utilize Eq.~\ref{eq:reconstruct} to reconstruct all old class prototypes in the latest subspace.  
 The complement process is training-free, making the learning process efficient.

\begin{table*}[t]
	\vspace{-5mm}
	\caption{\small Average and last performance comparison on seven datasets with {\bf ViT-B/16-IN21K} as the backbone.  `IN-R/A' stands for `ImageNet-R/A,' `ObjNet' stands for `ObjectNet,' and `OmniBench' stands for `OmniBenchmark.' 
		We report all compared methods with their source code.
		The best performance is shown in bold. All methods are implemented without using exemplars.
	}\label{tab:benchmark}
	\vspace{-3mm}
	\centering
	\resizebox{1.0\textwidth}{!}{%
		\begin{tabular}{@{}lccccccccc cccccccc}
			\toprule
			\multicolumn{1}{l}{\multirow{2}{*}{Method}} & 
			\multicolumn{2}{c}{CIFAR B0 Inc5} & \multicolumn{2}{c}{CUB B0 Inc10} 
			& \multicolumn{2}{c}{IN-R B0 Inc5}
			& \multicolumn{2}{c}{IN-A B0 Inc20}
			& \multicolumn{2}{c}{ObjNet B0 Inc10}
			& \multicolumn{2}{c}{OmniBench B0 Inc30}
			& \multicolumn{2}{c}{VTAB B0 Inc10} \\
			& {$\bar{\mathcal{A}}$} & ${\mathcal{A}_B}$  
			& {$\bar{\mathcal{A}}$} & ${\mathcal{A}_B}$
			& {$\bar{\mathcal{A}}$} & ${\mathcal{A}_B}$   
			& {$\bar{\mathcal{A}}$} & ${\mathcal{A}_B}$ 
			& {$\bar{\mathcal{A}}$} & ${\mathcal{A}_B}$ 
			& {$\bar{\mathcal{A}}$} & ${\mathcal{A}_B}$ 
			& {$\bar{\mathcal{A}}$} & ${\mathcal{A}_B}$ 
			\\
			\midrule
			Finetune	& 38.90 & 20.17 &26.08 & 13.96 &21.61 & 10.79 &24.28 & 14.51 & 19.14 & 8.73 & 23.61 & 10.57 & 34.95 & 21.25  \\
			Finetune Adapter~\cite{chenadaptformer} & 60.51 &49.32& 66.84 &52.99 & 47.59 &40.28 &45.41 &41.10 &50.22 &35.95 &62.32& 50.53 &48.91 & 45.12 \\
			LwF~\cite{li2017learning}& 46.29 & 41.07 &48.97 & 32.03  & 39.93 &26.47 &37.75 & 26.84 & 33.01 & 20.65 & 47.14 &33.95 & 40.48 & 27.54\\
			SDC~\cite{yu2020semantic} &68.21 &63.05 &  70.62 & 66.37 & 52.17 & 49.20 & 29.11 & 26.63 & 39.04 & 29.06 &60.94 & 50.28 &45.06 &22.50\\
			L2P~\cite{wang2022learning}   & 85.94 & 79.93 &67.05 & 56.25 & 66.53 & 59.22 &  49.39 & 41.71 &  63.78 & 52.19 &73.36 & 64.69 & 77.11 & 77.10\\
			DualPrompt~\cite{wang2022dualprompt}    &87.87 & 81.15& 77.47 & 66.54 & 63.31 & 55.22 & 53.71 & 41.67 & 59.27 & 49.33 & 73.92 & 65.52 & 83.36 & 81.23\\
			CODA-Prompt~\cite{smith2023coda} & 89.11 & 81.96 & 84.00 & 73.37 & 64.42 &55.08 & 53.54 & 42.73 & 66.07 &53.29 &77.03 &68.09 &83.90 &83.02\\
			SimpleCIL~\cite{zhou2023revisiting}   &  87.57 & 81.26 & 92.20 & 86.73 & 62.58 & 54.55 & 59.77 & 48.91 & 65.45 & 53.59 & 79.34 & 73.15 & 85.99 & 84.38\\
			\ADAM+ Finetune~\cite{zhou2023revisiting}   & 87.67 & 81.27 & 91.82 & 86.39 & 70.51 & 62.42 &   61.01 & 49.57 & 61.41 & 48.34 & 73.02 & 65.03 &  87.47 & 80.44\\
			\ADAM+ VPT-S~\cite{zhou2023revisiting}    &  90.43 & 84.57& 92.02 &86.51 & 66.63 &58.32 & 58.39 & 47.20& 64.54 & 52.53 & 79.63 & 73.68 & 87.15 &  85.36\\
			\ADAM + VPT-D~\cite{zhou2023revisiting} & 88.46 & 82.17 & 91.02 &84.99 & 68.79 & 60.48 & 58.48 & 48.52 & 67.83 & 54.65 &  81.05 &  74.47 & 86.59 & 83.06\\
			\ADAM + SSF~\cite{zhou2023revisiting}  & 87.78 & 81.98   & 91.72 &86.13& 68.94 & 60.60 &61.30 & 50.03 & 69.15 & 56.64 &  80.53 & 74.00 & 85.66 & 81.92\\
			\ADAM + Adapter~\cite{zhou2023revisiting} &   90.65 &  85.15 &92.21 &86.73 & 72.35 & 64.33 & 60.47 &49.37 &  67.18 & 55.24 &  80.75 & 74.37 &  85.95 & 84.35\\
			\midrule
			\name & \bf 91.51 & \bf 85.80 & \bf92.23& \bf86.81 & \bf78.31& \bf70.58 &\bf 65.34 & \bf55.04 &\bf 70.84 & \bf57.86 &\bf 81.11&\bf 74.85& \bf93.61&\bf 93.55\\
			\bottomrule
		\end{tabular}
	}
\end{table*}

\begin{figure*}
	\centering
	\begin{subfigure}{0.33\linewidth}
		\includegraphics[width=1\columnwidth]{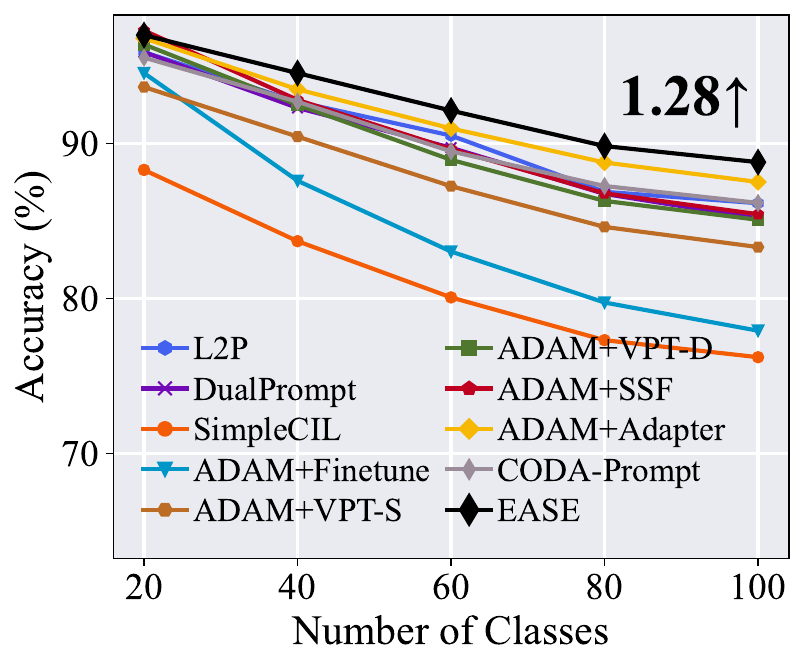}
		\caption{CIFAR B0 Inc20}
		\label{fig:benchmark-cifar}
	\end{subfigure}
	\hfill
	\begin{subfigure}{0.33\linewidth}
		\includegraphics[width=1\linewidth]{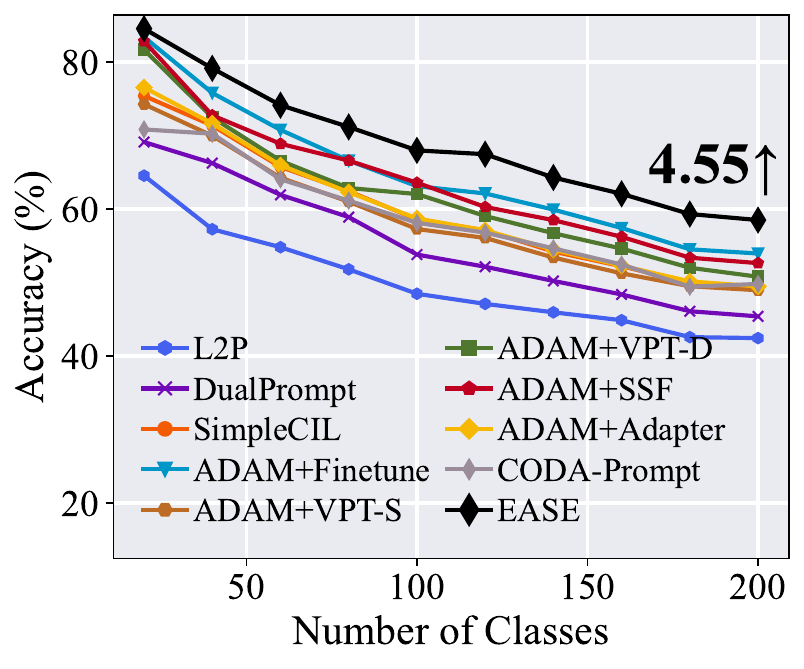}
		\caption{ImageNet-A B0 Inc20}
		\label{fig:benchmark-ina}
	\end{subfigure}
	\hfill
	\begin{subfigure}{0.33\linewidth}
		\includegraphics[width=1\linewidth]{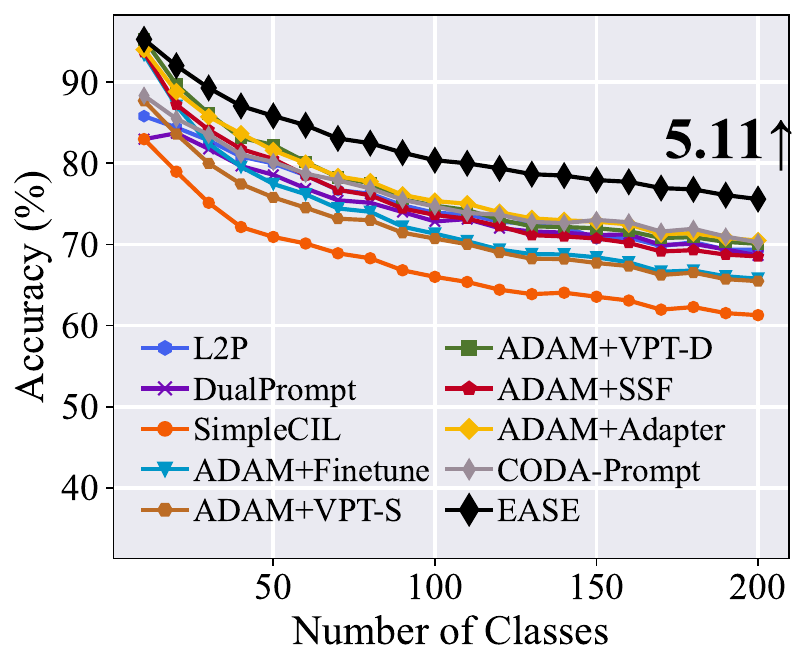}
		\caption{ImageNet-R B0 Inc10}
		\label{fig:benchmark-inr}
	\end{subfigure}
	\\
	\begin{subfigure}{0.33\linewidth}
		\includegraphics[width=1\linewidth]{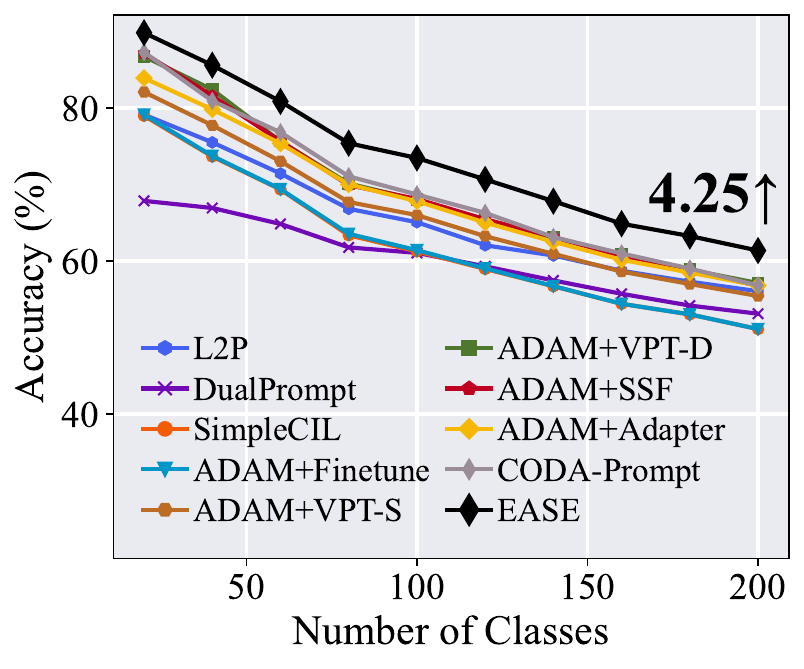}
		\caption{ObjectNet B0 Inc20}
		\label{fig:benchmark-obj}
	\end{subfigure}
	\hfill
	\begin{subfigure}{0.33\linewidth}
		\includegraphics[width=1\linewidth]{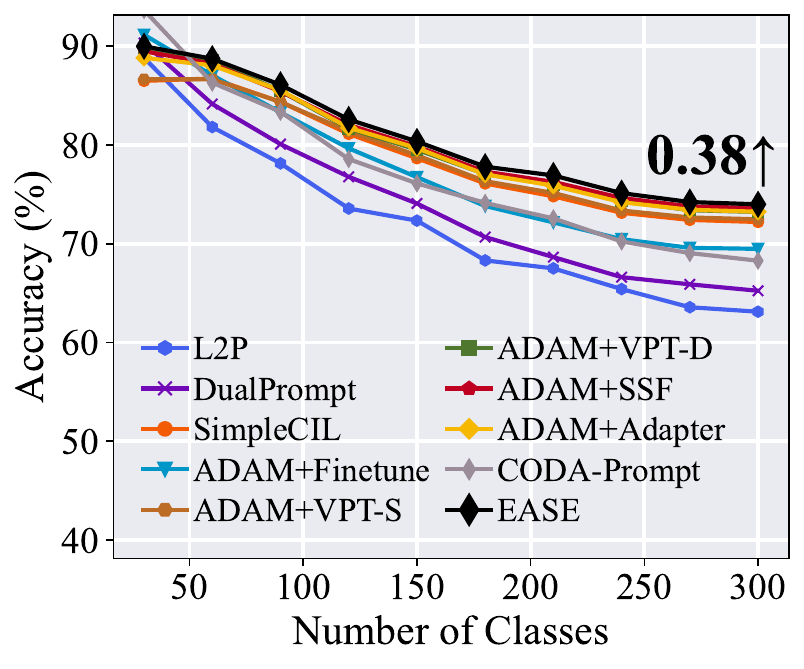}
		\caption{Omnibenchmark B0 Inc30}
		\label{fig:benchmark-omni}
	\end{subfigure}
	\hfill
	\begin{subfigure}{0.33\linewidth}
		\includegraphics[width=1\columnwidth]{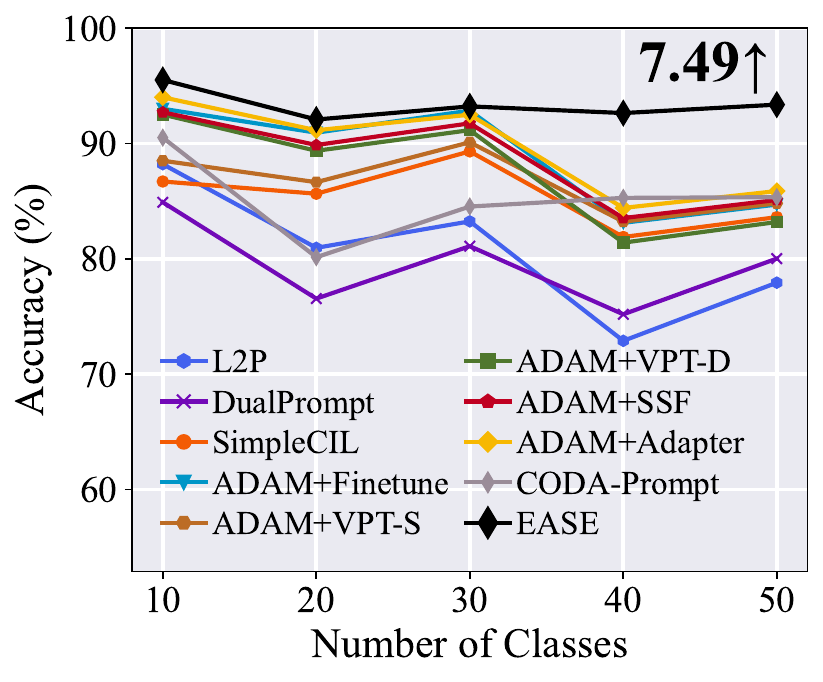}
		\caption{VTAB B0 Inc10}
		\label{fig:benchmark-vtab}
	\end{subfigure}
	\vspace{-3mm}
	\caption{\small Performance curve of different methods under different settings. All methods are initialized with {\bf ViT-B/16-IN1K}. We annotate the relative improvement of \name above the runner-up method with numerical numbers at the last incremental stage. }
	\vspace{-3mm}
	\label{fig:benchmark}
\end{figure*}

\subsection{Subspace Ensemble via Subspace Reweight}

So far, we have introduced subspace expansion with new adapters and prototype complement to restore old class prototypes. 
After adapter expansion and prototype complement, we can get a full classifier (prototype matrix) as:
\begin{equation} \label{eq:get_w}
	\left[\begin{array}{cccc}
		\mathbf{P}_{1,1} &\hat{\mathbf{P}}_{1,2} & \cdots & \hat{\mathbf{P}}_{1,B} \\
		\mathbf{P}_{2,1} & \mathbf{P}_{2,2} & \cdots & \hat{\mathbf{P}}_{2,B} \\
		\vdots & \vdots & \ddots & \vdots \\
		\mathbf{P}_{B,1} & \mathbf{P}_{B,1} & \cdots & \mathbf{P}_{B,B}
	\end{array}\right] \,.
\end{equation}
Note that items above the main diagonal are estimated via Eq.~\ref{eq:reconstruct}. 
 During inference, the logit of task $b$ is calculated by:
\begin{equation} \textstyle \label{eq:pred}
	[\mathbf{P}_{b,1},\mathbf{P}_{b,2},\cdots,\mathbf{P}_{b,B}]^\top \Phi(\x)=\sum_i \mathbf{P}_{b,i}^\top \phi(\x;\A_i) \,,
\end{equation}
which equals the {\em ensemble} of multiple (prototype-embedding) matching logit in different subspaces.
Among the items in Eq.~\ref{eq:pred}, only adapter $\A_b$ is especially learned to extract task-specific features for the $b$-th task. Hence, we think these prototypes are more suitable for classifying the corresponding task and should take a greater part in the final inference. Hence, we transform
 Eq.~\ref{eq:pred} by assigning higher weights to the matching subspace:
\begin{equation} \textstyle \label{eq:pred_ours}
 \mathbf{P}_{b,b}^\top \phi(\x;\A_b) + \alpha
\sum_{i\neq b} \mathbf{P}_{b,i}^\top \phi(\x;\A_i) \,,
\end{equation}
where $\alpha$ is the trade-off parameter, which is set to $0.1$ in our experiments. 
Reweighting the logits enables us to highlight the contributions of core features in the decision.

\noindent\textbf{Summary of \mame}: We summarize the training pipeline of \name in the supplementary. We initialize and train an adapter for each incoming task to encode the task-specific information. 
Afterward, we extract the prototypes of the current dataset for all adapters and synthesize the prototypes of former classes. Finally, we construct the full classifier and reweight the logit for prediction.
Since we are using the prototype-based classifier for inference, the classifier $W$ in Eq.~\ref{eq:adapter} will be dropped after each learning stage.

%% file: experiments.tex
\section{Experiments}

In this section, we conduct experiments on seven benchmark datasets and compare \name to other state-of-the-art algorithms to show the incremental learning ability. Additionally, we provide an ablation study and parameter analysis to investigate the robustness of our proposed method. We also analyze the effect of prototype synthesis and provide visualization to show \mame's effectiveness. More experimental results can be found in the supplementary.

\subsection{Implementation Details}
\noindent {\bf Dataset}: Since pre-trained models may possess extensive knowledge of upstream tasks, we follow~\cite{zhou2023revisiting,wang2022learning} to evaluate the performance on CIFAR100~\cite{krizhevsky2009learning}, CUB200~\cite{WahCUB2002011}, ImageNet-R~\cite{hendrycks2021many}, ImageNet-A~\cite{hendrycks2021natural}, ObjectNet~\cite{barbu2019objectnet}, Omnibenchmark~\cite{zhang2022benchmarking} and VTAB~\cite{zhai2019large}. These datasets contain typical CIL benchmarks and out-of-distribution datasets that have {\em large domain gap} with ImageNet (\ie, the pre-trained dataset). There are 50 classes in VTAB, 100 classes in CIFAR100, 200 classes in CUB, ImageNet-R, ImageNet-A, ObjectNet, and 300 classes in OmniBenchmark. More details are reported in the supplementary.

\begin{table}[t]
	\vspace{-3mm}
	\caption{ Comparison to traditional exemplar-based CIL methods. \name does not use any exemplars. All methods are based on the same pre-trained model ({\bf ViT-B/16-IN21K}).
	}  
	\label{tab:benchmark-typicalmethods}
	\centering
	\vspace{-3mm}
	\resizebox{0.98\columnwidth}{!}{%
		\begin{tabular}{@{}lccccccccc }
			\toprule
			\multicolumn{1}{l}{\multirow{2}{*}{Method}} &
			\multicolumn{1}{l}{\multirow{2}{*}{Exemplars}} & 
			\multicolumn{2}{c}{ImageNet-R B0 Inc20} & \multicolumn{2}{c}{CIFAR B0 Inc10}  \\
			& & {$\bar{\mathcal{A}}$} & ${\mathcal{A}_B}$  
			& {$\bar{\mathcal{A}}$} & ${\mathcal{A}_B}$
			\\
			\midrule
			iCaRL~\cite{rebuffi2017icarl}& 20 / class & 72.42&60.67 & 82.46& 73.87 \\
			DER~\cite{yan2021dynamically} &20 / class  & 80.48& 74.32 & 86.04& 77.93\\
			FOSTER~\cite{wang2022foster} &20 / class   & 81.34&74.48 &89.87& 84.91\\
			MEMO~\cite{zhou2022model} &20 / class &74.80& 66.62 & 84.08& 75.79\\
			\midrule
			\name  &\bf 0 & \bf 81.73& \bf 76.17  & \bf 92.35 & \bf 87.76\\
			\bottomrule
		\end{tabular}
	}
	\vspace{-3mm}
\end{table}

\begin{figure}
	\centering
	\begin{subfigure}{0.49\linewidth}
		\includegraphics[width=1\columnwidth]{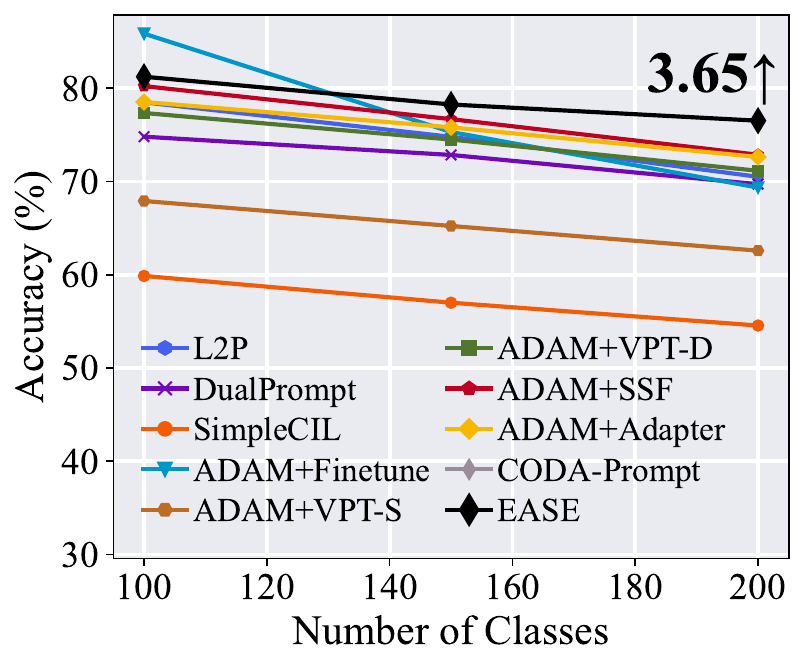}
		\caption{ImageNet-R B100 Inc50}
		\label{fig:benchmark-large-base-a}
	\end{subfigure}
	\hfill
	\begin{subfigure}{0.49\linewidth}
		\includegraphics[width=1\linewidth]{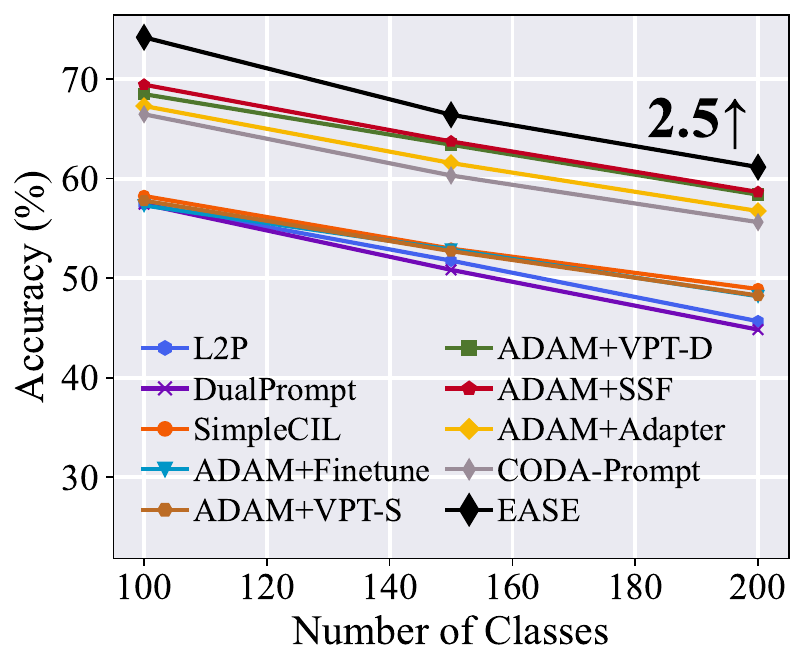}
		\caption{ImageNet-A B100 Inc50}
		\label{fig:benchmark-large-base-b}
	\end{subfigure}
	\vspace{-2mm}
	\caption{\small Experimental results with large base classes. All methods are based on the same pre-trained model ({\bf ViT-B/16-IN21K})}
	\vspace{-4mm}
	\label{fig:benchmark-large-base}
\end{figure}

\noindent {\bf Dataset split:} Following the benchmark setting~\cite{rebuffi2017icarl,wang2022learning}, we use `B-$m$ Inc-$n$' to denote the class split.
$m$ indicates the number of classes in the first stage, and $n$ represents that of every incremental stage. For all compared methods, we follow~\cite{rebuffi2017icarl} to randomly shuffle class orders with random seed 1993 before data split. We keep the training and testing set the same as~\cite{zhou2023revisiting} for all methods for a fair comparison.

\noindent {\bf Comparison methods:} We choose state-of-the-art PTM-based CIL methods for comparison, \ie, L2P~\cite{wang2022learning}, DualPrompt~\cite{wang2022dualprompt}, CODA-Prompt~\cite{smith2023coda}, SimpleCIL~\cite{zhou2023revisiting} and ADAM~\cite{zhou2023revisiting}. Additionally, we also compare our method to typical CIL methods by equipping them with the {\em same} PTM, \eg, LwF~\cite{li2017learning}, SDC~\cite{yu2020semantic}, iCaRL~\cite{rebuffi2017icarl}, DER~\cite{yan2021dynamically}, FOSTER~\cite{wang2022foster} and MEMO~\cite{zhou2022model}.
We report the baseline method, which sequentially finetunes the PTM as Finetune. We implement all methods with the {\bf same PTM}.

\noindent {\bf Training details:}
We run experiments on NVIDIA 4090 and reproduce other compared methods with PyTorch~\cite{paszke2019pytorch} and Pilot~\cite{sun2023pilot}. Following~\cite{wang2022learning,zhou2023revisiting}, we consider two representative models, \ie, ViT-B/16-IN21K and ViT-B/16-IN1K as the pre-trained model. They are obtained by pre-training on ImageNet21K, while the latter is further finetuned with ImageNet1K. In \mame, we train the model using SGD optimizer, with a batch size of $48$ for $20$ epochs. The learning rate decays from $0.01$ with cosine annealing. We set the projection dim $r$ in the adapter to $16$ and the trade-off parameter $\alpha$ to $0.1$.

\noindent {\bf Evaluation metric:} Following the benchmark protocol~\cite{rebuffi2017icarl}, we use $\mathcal{A}_b$ to represent the model's accuracy after the $b$-th stage. Specifically, we adopt $\mathcal{A}_B$ (the performance after the last stage) and $\bar{\mathcal{A}}=\frac{1}{B}\sum_{b=1}^{B}\mathcal{A}_b$ (average performance along incremental stages) as measurements.

\begin{figure}[t]
	\vspace{-3mm}
	\begin{center}
		{\includegraphics[width=0.8\columnwidth]{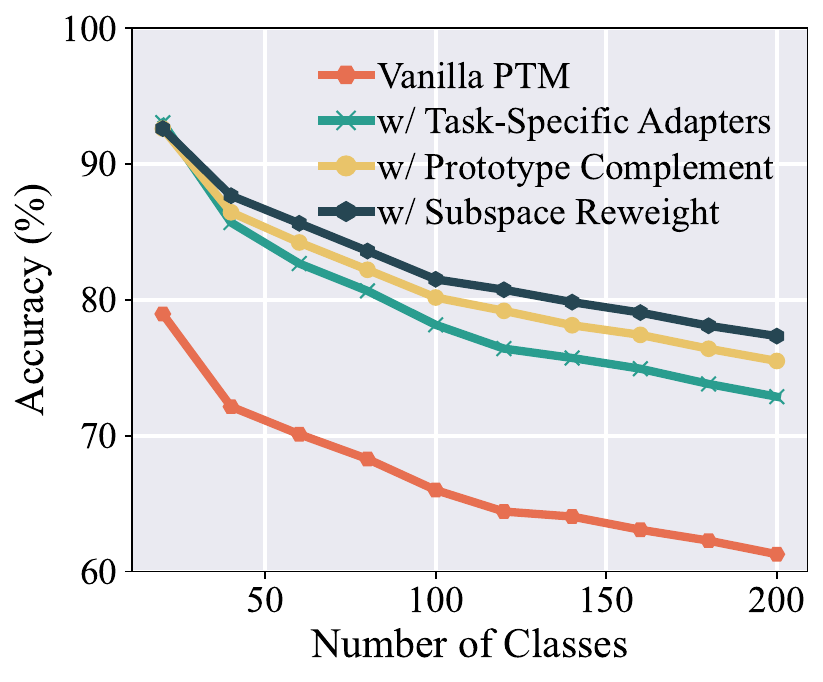}}
	\end{center}
	\vspace{-6mm}
	\caption{\small  Ablation Study of different components in \mame. We find every component in \name can improve the performance.
	}
	\vspace{-5mm}
	\label{fig:ablation}
\end{figure}

\subsection{Benchmark Comparison}

In this section, we compare \name to other state-of-the-art methods on seven benchmark datasets and different backbone weights. 
Table~\ref{tab:benchmark} reports the comparison of different methods with ViT-B/16-IN21K. We can infer that \name achieves the best performance among all seven benchmarks, substantially outperforming the current SOTA methods, \ie, CODA-Prompt, and ADAM. We also report the incremental performance trend of different methods in Figure~\ref{fig:benchmark} with ViT-B/16-IN1K. As annotated at the end of each image, we find \name outperforms the runner-up method by 4\%$\sim$7.5\% on ImageNet-R/A, ObjectNet, and VTAB. 

Apart from the B0 settings in Table~\ref{tab:benchmark} and Figure~\ref{fig:benchmark}, we also conduct experiments with vase base classes. As shown in Figure~\ref{fig:benchmark-large-base}, \name still works competitively given various data split settings. Additionally, we also compare \name to traditional CIL methods by implementing them based on the same pre-trained model in Table~\ref{tab:benchmark-typicalmethods}. It must be noted that traditional CIL methods require saving exemplars to recover former knowledge, while ours do not. We follow~\cite{rebuffi2017icarl} to set the exemplar number to 20 per class for these methods.
Surprisingly, we find \name still works competitively in comparison to these exemplar-based methods.

Finally, we investigate the parameter number of different methods and report the parameter-performance comparison on ImageNet-R B100 Inc50 in Figure~\ref{figure:parameter-size}. As shown in the figure, \name uses the same scale of parameters as other prompt-based methods, \eg, L2P and DualPrompt, while achieving the best performance among all competitors. Extensive experiments validate the effectiveness of \mame.

\subsection{Ablation Study}
In this section, we conduct an ablation study to investigate the effectiveness of each component in \mame. Specifically, we report the incremental performance of different variations on ImageNet-R B0 Inc20 in Figure~\ref{fig:ablation}. In the figure, `{\bf Vanilla PTM}' denotes classifying with prototype classifier of the pre-trained image encoder, which stands for the baseline. 
To enhance feature diversity, we aim to equip the PTM with expandable adapters (Eq.~\ref{eq:expand_adapter}). Since we do not have exemplars, we report the performance of `{\bf w/ Task-Specific Adapters}' by only using the diagonal components in Eq.~\ref{eq:get_w}. When comparing it to `Vanilla PTM,' we find although a pre-trained model possesses generalizable features, the adaptation to downstream tasks to extract task-specific features is also an essential step in CIL.
Furthermore, we can complete the classifier by semantic mapping (Eq.~\ref{eq:reconstruct}) and use a full classifier instead of diagonal components for classification. We denote such format as `{\bf w/ Prototype Complement}.' As shown in the figure, prototype complement further improves the performance, indicating that cross-task semantic information from other tasks can help the inference. 
Finally, we adjust the logit with Eq.~\ref{eq:pred_ours} by reweighting the importance of different components (denoted as `{\bf w/ Subspace Reweight}'), which further improves the performance. Ablations verify that every component in \name boosts the CIL performance.

\begin{figure}
	\vspace{-3mm}
	\centering
	\begin{subfigure}{0.49\linewidth}
		\includegraphics[width=1\columnwidth]{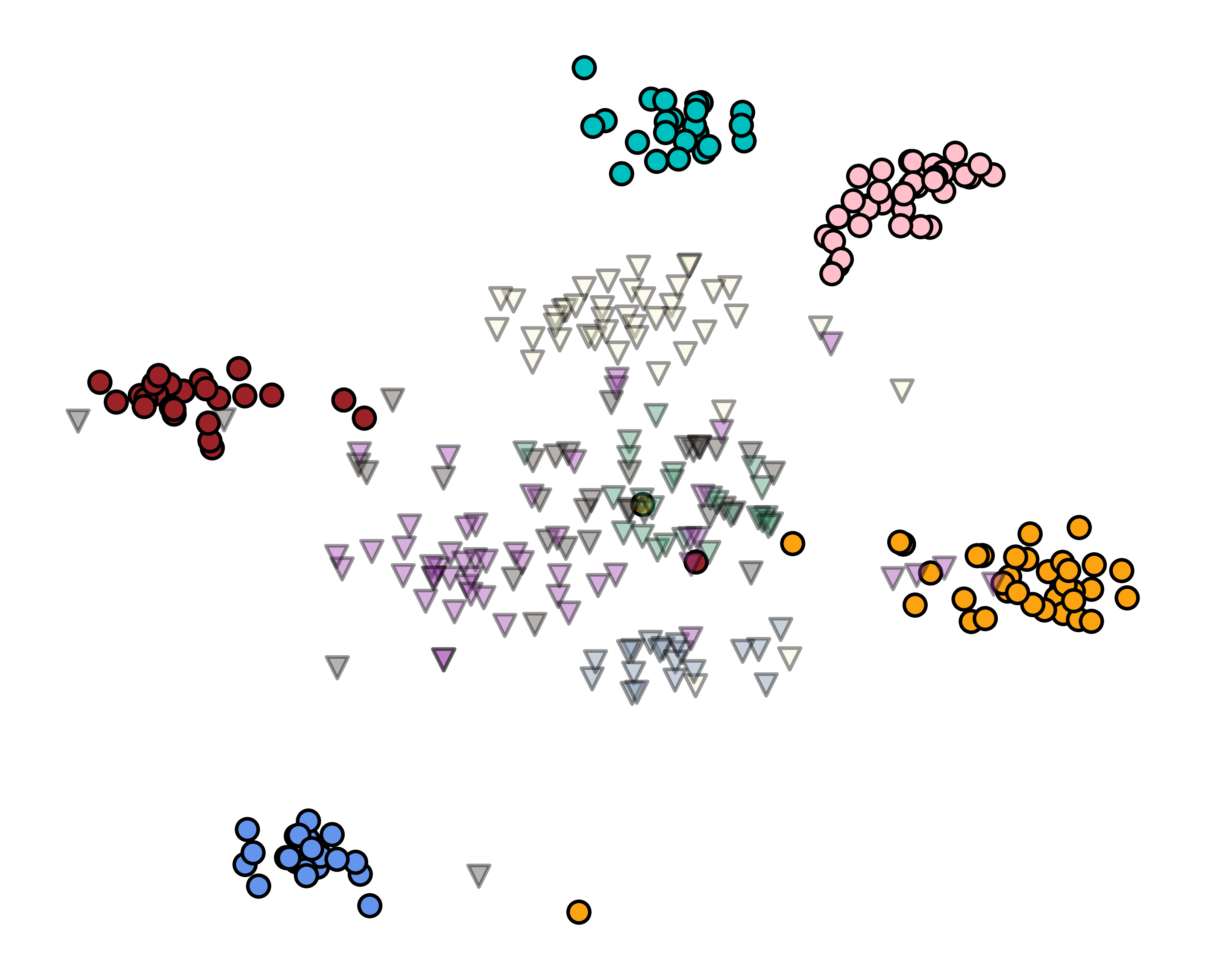}
		\caption{Subspace of $\A_1$}
		\label{fig:vis-a}
	\end{subfigure}
	\hfill
	\begin{subfigure}{0.49\linewidth}
		\includegraphics[width=1\linewidth]{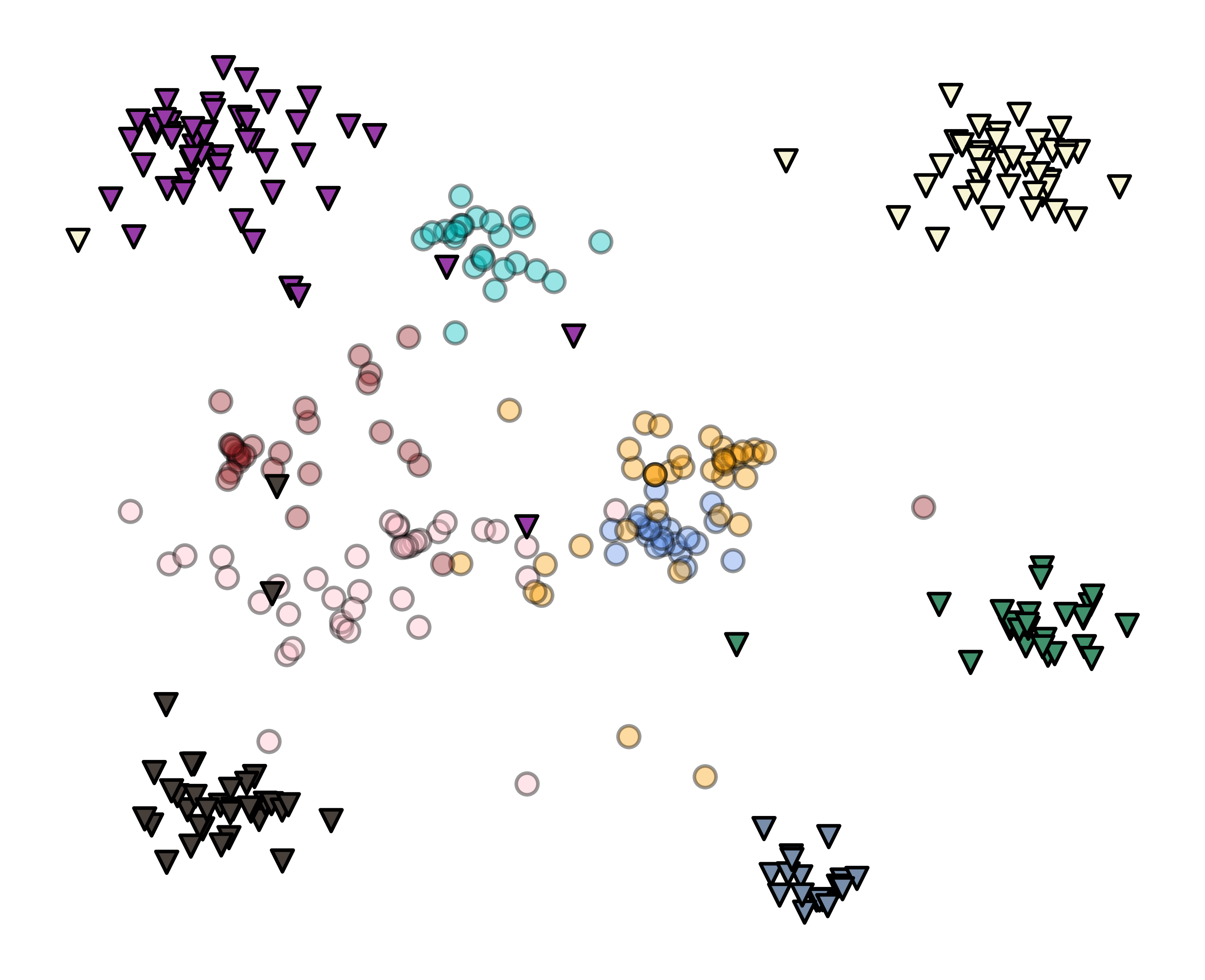}
		\caption{Subspace of $\A_2$}
		\label{fig:vis-b}
	\end{subfigure}
	\vspace{-2mm}
	\caption{\small t-SNE~\cite{van2008visualizing} visualizations of different adapters' subspaces, which are learned to discriminate the corresponding task.}
	\vspace{-3mm}
	\label{fig:vis}
\end{figure}

\subsection{Further Analysis}

\noindent\textbf{Visualizations}: In this paper, we expect different adapters to learn task-specific features. To verify this hypothesis, we conduct experiments with ImageNet-R B0 Inc5 and visualize the embeddings in different adapter spaces in Figure~\ref{fig:vis} using t-SNE~\cite{van2008visualizing}. We consider two incremental stages (each containing five classes) and learn two adapters $\A_1, \A_2$ for these tasks. We represent classes of the first task with dots and classes of the second task with triangles. As shown in Figure~\ref{fig:vis-a}, in adapter $\A_1$'s embedding space, classes of the first task (dots) are clearly separated, while classes of the second task (triangles) are not. We can observe a similar phenomenon in Figure~\ref{fig:vis-b}, where adapter $\A_2$ can discriminate classes in the second task. Hence, we should mainly resort to the adapter to classify classes of the corresponding task, as formulated in Eq.~\ref{eq:pred_ours}.

\begin{figure}
	\vspace{-3mm}
	\centering
	\begin{subfigure}{0.49\linewidth}
		\includegraphics[width=1\columnwidth]{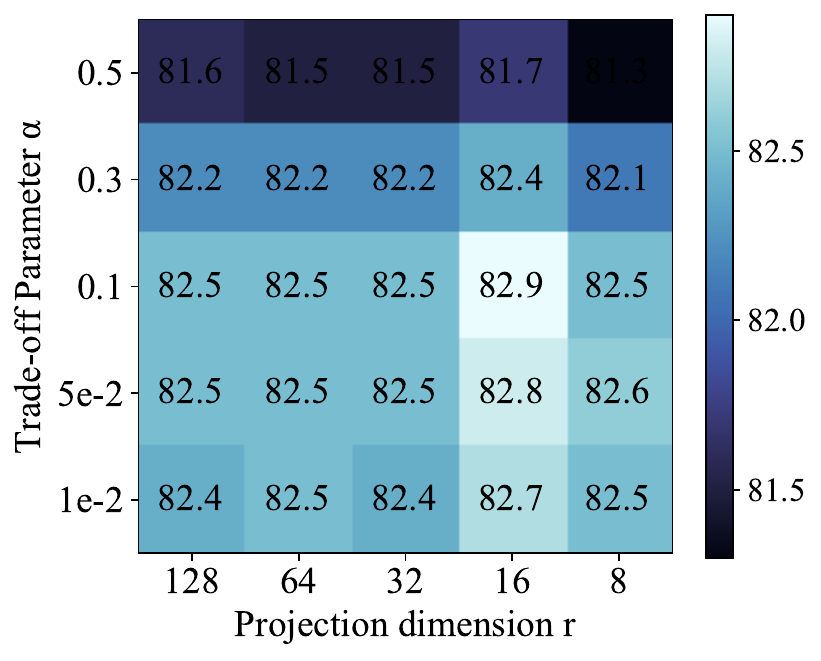}
		\caption{Robustness of hyperparameters}
		\label{fig:para-robust}
	\end{subfigure}
	\hfill
	\begin{subfigure}{0.49\linewidth}
		\includegraphics[width=1\linewidth]{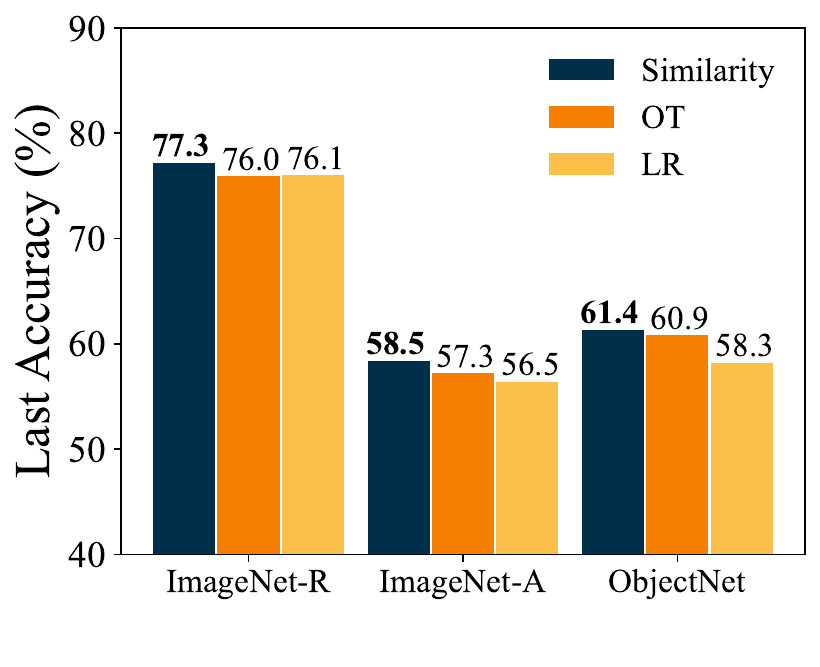}
		\caption{Variations of Eq.~\ref{eq:reconstruct} }
		\label{fig:prototype-complement}
	\end{subfigure}
		\vspace{-2mm}
	\caption{\small Further analysis on parameter robustness and prototype complement strategy. }
		\vspace{-3mm}
	\label{fig:further-analysis}
\end{figure}

\noindent\textbf{Parameter robustness}: There are two hyperparameters in \mame, \ie, the projection dim $r$ in the adapter and the trade-off parameter $\alpha$ in Eq.~\ref{eq:pred_ours}. We conduct experiments on ImageNet-R B0 Inc20 to investigate the robustness by changing these parameters. Specifically, we choose $r$ among $\{8, 16, 32, 64, 128\}$, and $\alpha$ among $\{0.01, 0.05, 0.1, 0.3, 0.5\}$. We report the average performance in Figure~\ref{fig:para-robust}. As shown in the figure, the performance is robust with the change of these parameters, and we suggest $r=16, \alpha=0.1$ as default for other datasets.

\noindent\textbf{Prototype complement}: Apart from similarity-based mapping in Eq.~\ref{eq:reconstruct}, there are other ways to learn the mapping and complete the prototype matrix, \eg, Linear Regression (LR) and Optimal Transport (OT)~\cite{kantorovich1960mathematical,ye2018rectify}. Hence, we also compare the similarity-based complement to these variations in Figure~\ref{fig:prototype-complement}. With other settings the same, we find the current complement strategy the best among these variations.

%% file: conclusion.tex
\section{Conclusion} \label{sec:conclusion}

Incremental learning is a desired ability of real-world learning systems. This paper proposes expandable subspace ensemble (\mame) for class-incremental learning with a pre-trained model. Specifically, we equip a PTM with diverse subspaces through lightweight adapters. Aggregating historical features enables the model to extract holistic embeddings without forgetting. Besides, we utilize semantic information to synthesize the prototypes of former classes in latter subspaces without the help of exemplars. Extensive experiments verify \mame's effectiveness.
\\\noindent\textbf{Limitations and future works:} 
Although adapters are lightweight modules that only consume limited parameters (0.3\% of the total backbone), possible limitations include the extra model size for saving these adapters. Future works include designing algorithms to compress adapters.

\section*{Acknowledgments}
This work is partially supported by National Key R\&D Program of China (2022ZD0114805),
NSFC (62376118, 62006112, 62250069, 61921006), Collaborative Innovation Center of Novel Software
Technology and Industrialization.

%% file: supp_material.tex
\section{Further Ablations} \label{sec:supp_further_ablation}
In this section, we conduct further analysis on \mame's components to investigate their effectiveness, \eg, semantic-guided mapping and adapter-spanned subspaces. We also include the comparison about random seeds, running time, and the results of the upper bound.

\subsection{Prototype-Prototype Similarity VS. Prototype-Instance Similarity}

In the main paper, we formulate the prototype complement task as: given two subspaces (old and new) and two class sets (old and new), the target is to estimate old class prototypes in the new subspace $\hat{\mathbf{P}}_{o,n}$ using $\mathbf{P}_{o,o}$, $\mathbf{P}_{n,o}$, $\mathbf{P}_{n,n}$.
Among them, $\mathbf{P}_{o,o}$ and $\mathbf{P}_{n,o}$ represent prototypes of old and new classes in the old subspace (which we call co-occurrence space), and $\mathbf{P}_{n,n}$ represents new classes prototypes in the new subspace. 

\begin{figure}
	\centering
	\begin{subfigure}{0.8\linewidth}
		\includegraphics[width=1\columnwidth]{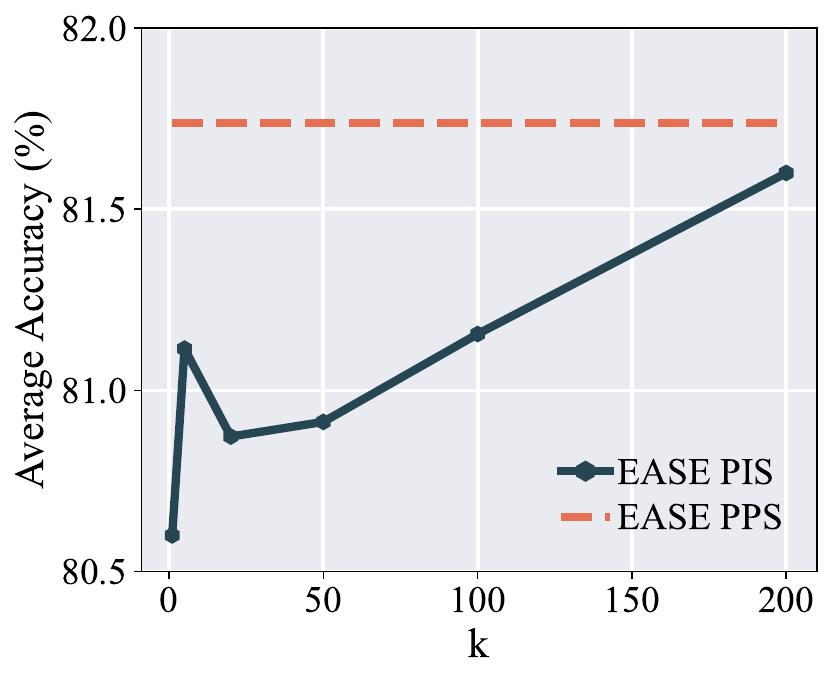}
		\caption{ImageNet-R B0 Inc20}
	\end{subfigure}
	\begin{subfigure}{0.8\linewidth}
		\includegraphics[width=1\linewidth]{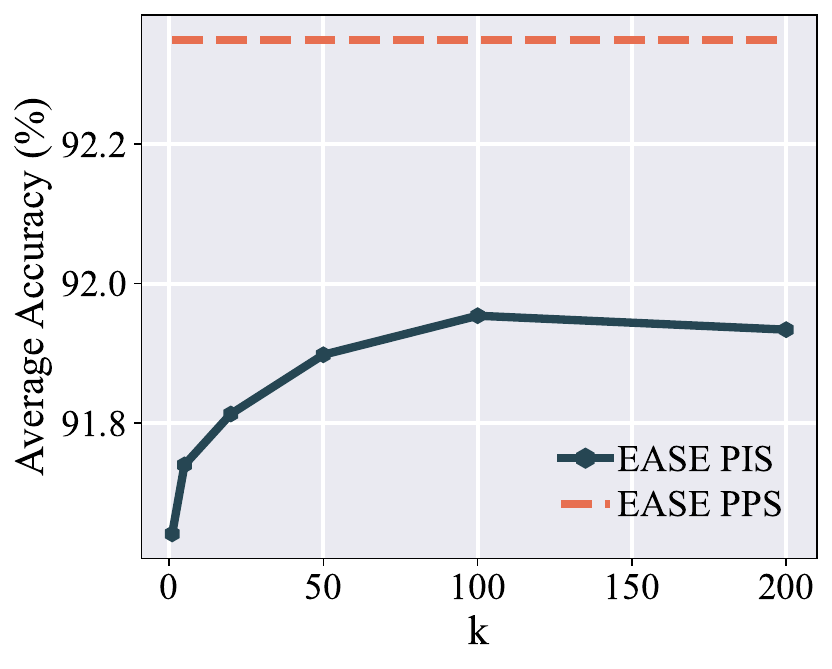}
		\caption{CIFAR100 B0 Inc10}
	\end{subfigure}
	\caption{ Experimental results on different similarity calculation methods. {\bf Using prototype-prototype similarity shows better performance than prototype-instance similarity. }}
	\label{fig:supp-p2i}
\end{figure}

During the complement process, we construct a class-wise similarity matrix in the old subspace:
\begin{equation} \label{eq:supp-sim}
	\text{Sim}_{i,j}=\frac{\mathbf{P}_{o,o}[i]}{\|\mathbf{P}_{o,o}[i]\|_2} \frac{\mathbf{P}_{n,o}[j]^\top}{\|\mathbf{P}_{n,o}[j]\|_2} \,,
\end{equation}
and then utilize it to reconstruct prototypes via class-wise similarity in the new subspace:
\begin{equation} \label{eq:supp-reconstruct} 
	\hat{\mathbf{P}}_{o,n}[i]=\sum_j  \text{Sim}_{i,j} \times \mathbf{P}_{n,n}[j] \,.
\end{equation} 

However, since we have the current dataset $\D^b$ in hand, apart from class-wise similarity, we can also measure the similarity of old class prototypes and new class instances.
\begin{equation} \label{eq:sim-proto2ins}
	\text{Sim}_{i,j}=\frac{\mathbf{P}_{n,o}[i]}{\|\mathbf{P}_{n,o}[i]\|_2} \frac{\phi(\x_j;\A_{old})^\top}{\|\phi(\x_j;\A_{old})\|_2} \,.
\end{equation}
Different from prototype to prototype similarity in Eq.~\ref{eq:supp-sim}, Eq.~\ref{eq:sim-proto2ins} measures the similarity of an old class prototype to a new class instance in the same subspace. In the implementation, we can choose $\x_j$ in a subset containing $k$ instances and obtain a similarity matrix of $|Y_{old}|\times k$. The choice of these $k$ instances is based on the relative similarity. 
Similar to the reconstruction process in Eq.~\ref{eq:supp-reconstruct}, we can build the prototype complement process via:
\begin{equation} \label{eq:reconstruct2}
	\hat{\mathbf{P}}_{o,n}[i]=\sum_j  \text{Sim}_{i,j} \times \phi(\x_j;\A_{new}) \,.
\end{equation}	
We call the prototype-instance similarity-based complement process in Eq.~\ref{eq:reconstruct2} as PIS (prototype-instance similarity) while calling the prototype-prototype similarity-based complement process in Eq.~\ref{eq:supp-reconstruct} as PPS (prototype-prototype similarity). In this section, we conduct experiments on CIFAR100 and ImageNet-R to compare these variations. We utilize ViT-B/16-IN21K as the backbone and keep other settings the same. We choose $k$ in PIS among $\{1,5,20,50,100,200\}$.

We report the experimental results in Figure~\ref{fig:supp-p2i}. As shown in the figure, utilizing more instances (\ie, with larger $k$) shows better performance. However, we find using prototype-instance similarity less effective than using prototype-prototype similarity, even consuming more resources.

\begin{figure}
	\centering
	\begin{subfigure}{0.8\linewidth}
		\includegraphics[width=1\columnwidth]{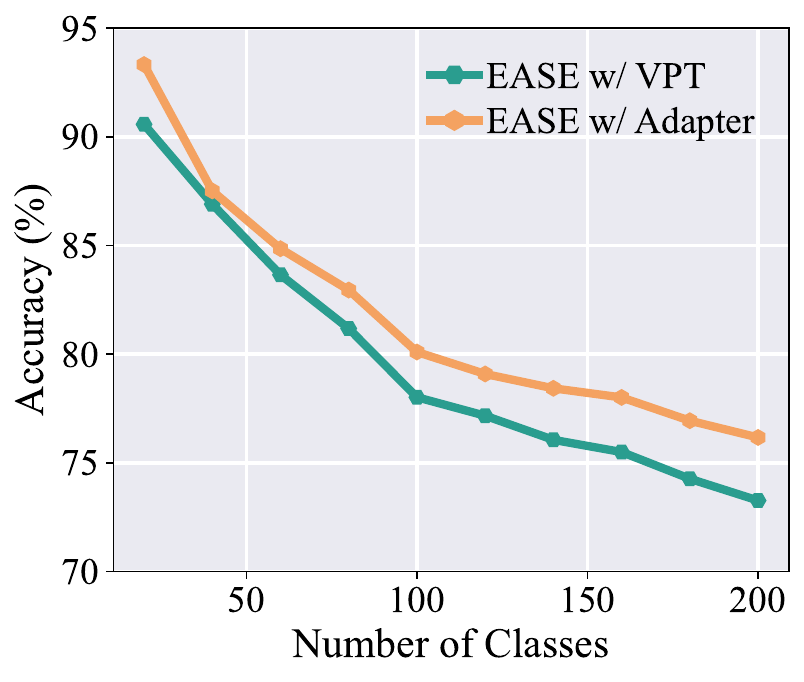}
		\caption{ImageNet-R B0 Inc20}
	\end{subfigure}
	\begin{subfigure}{0.8\linewidth}
		\includegraphics[width=1\linewidth]{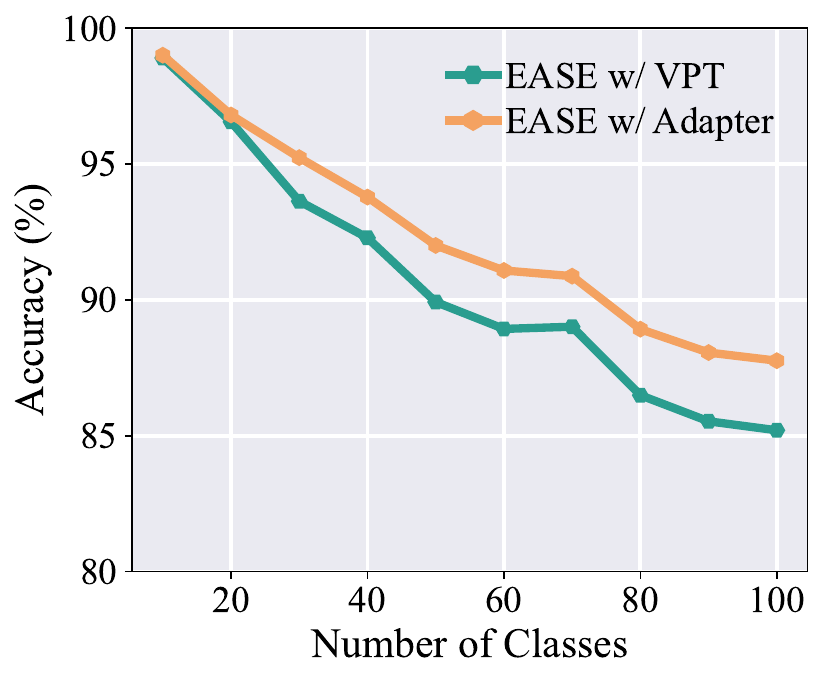}
		\caption{CIFAR100 B0 Inc10}
	\end{subfigure}
	\caption{ Experimental results on different subspace tuning methods. {\bf Using adapter tuning shows better performance than VPT. }}
	\label{fig:supp-vpt-vs-adapter}
\end{figure}

\subsection{Adapter VS. VPT}

In the main paper, we build task-specific subspaces via adapter tuning~\cite{chenadaptformer}. However, apart from adapter tuning, there are other ways to tune the pre-trained model in a parameter-efficient manner, \eg, visual prompt tuning~\cite{jia2022visual} (VPT). In this section, we combine our method with different subspace build techniques and combine \name with adapter and VPT, respectively. We conduct experiments on CIFAR100 and ImageNet-R. We keep other settings the same and only change the way of subspace building, and report results in Figure~\ref{fig:supp-vpt-vs-adapter}.

As we can infer from the figure, using adapters to build subspaces shows better performance than using VPT, outperforming it by $2-3$\% on these datasets. The main reason lies in the difference between VPT and adapter, where adapter tuning shows to be a stronger tuning method for pre-trained models. Hence, we choose adapter tuning as the way to build subspaces in \mame.

\subsection{Comparison to Upper bound}

In the main paper, we conduct inference using the completed prototypes. However, if we can save a subset of exemplars $\mathcal{E}$ from former classes, we do not need to complete former class prototypes and can directly calculate them via:
\begin{equation}  \label{eq:supp-prototype}
	{\bm p}_{i,b}= \frac{1}{N} \sum_{j=1}^{|\mathcal{E}|}\mathbb{I}(y_j=i)\phi(\x_j;\A_b) \,.
\end{equation} 
We denote such a calculation process as the upper bound since the prototypes calculated via Eq.~\ref{eq:supp-prototype} are accurate estimations of the class center. In this section, we compare \name to upper bound to show its effectiveness and report the results in Table~\ref{tab:supp-upperbound}.

As we can infer from the table, \name shows competitive performance to the upper bound, achieving almost the same results without using any exemplars. Results verify the effectiveness of using semantic information to conduct prototype complement.

\begin{table}[t]
	\caption{ Comparison to exemplar-based upper bound. {\bf \name does not use any exemplars while showing competitive performance}.
	}  
	\label{tab:supp-upperbound}
	\centering
	\resizebox{0.98\columnwidth}{!}{%
		\begin{tabular}{@{}lccccccccc }
			\toprule
			\multicolumn{1}{l}{\multirow{2}{*}{Method}} &
			\multicolumn{1}{l}{\multirow{2}{*}{Exemplars}} & 
			\multicolumn{2}{c}{ImageNet-R B0 Inc20} & \multicolumn{2}{c}{CIFAR B0 Inc10}  \\
			& & {$\bar{\mathcal{A}}$} & ${\mathcal{A}_B}$  
			& {$\bar{\mathcal{A}}$} & ${\mathcal{A}_B}$
			\\
			\midrule
			Upper Bound &20 / class &\bf 81.73& 76.08 & 92.32& \bf 87.79\\
			\midrule
			\name  &\bf 0 & \bf 81.73& \bf 76.17  & \bf 92.35 &  87.76\\
			\bottomrule
		\end{tabular}
	}
\end{table}

\begin{figure}
	\centering
	\includegraphics[width=0.8\columnwidth]{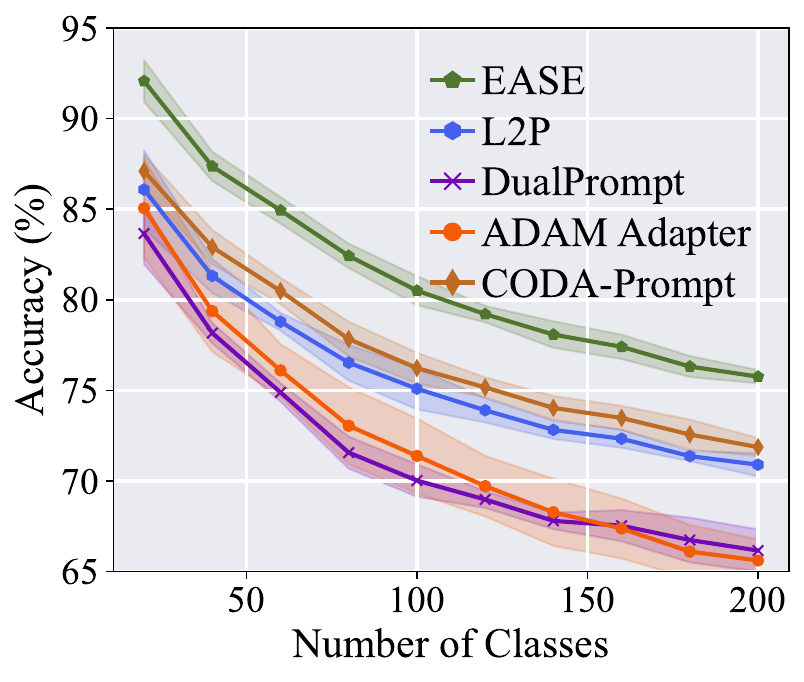}
	\caption{Results on ImageNet-R B0 Inc20 with multiple runs.  {\bf \name consistently outperforms other methods by a substantial margin. }}
	\label{fig:supp-multiple}
\end{figure}

\subsection{Multiple Runs}

In the main paper, we conduct experiments on different datasets and follow~\cite{rebuffi2017icarl} to shuffle class orders with random seed 1993. In this section, we also run the experiments multiple times using different random seeds, \ie, \{1993,1994,1995,1996,1997\}. Hence, we can obtain five incremental results of different methods and report the mean and standard variance in Figure~\ref{fig:supp-multiple}.

As we can infer from the figure, \name consistently outperforms other methods by a substantial margin given various random seeds.

\subsection{Running Time Comparison}

\begin{figure}
	\centering
		\includegraphics[width=0.9\columnwidth]{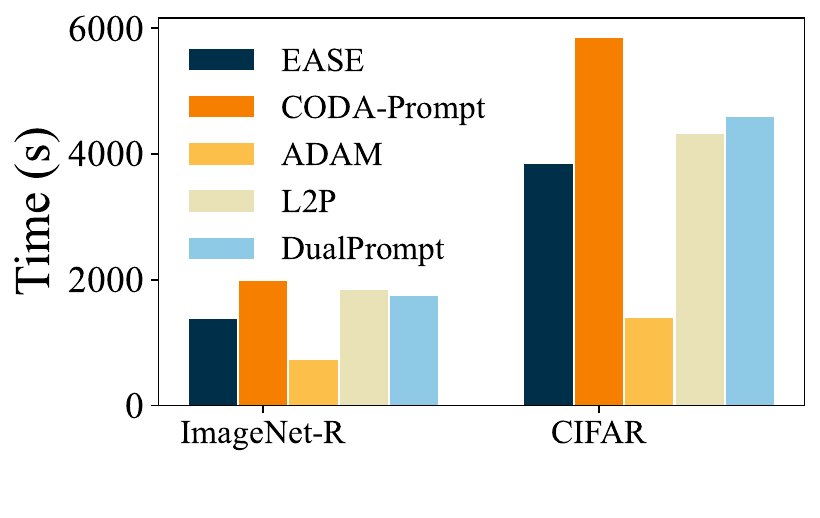}
		\caption{Running time comparison of different methods. {\bf \name utilizes less running time than CODA-Prompt, L2P, and DualPrompt while having better performance. }}
	\label{fig:supp-running-time}
\end{figure}

In this section, we report the running time comparison of different methods. We utilize a single NVIDIA 4090 GPU to run the experiments and report the results in Figure~\ref{fig:supp-running-time}. As we can infer from the figure, \name requires less running time than CODA-Prompt, L2P, and DualPrompt, while having the best performance.
Experimental results verify the effectiveness of \mame.

\begin{figure*}
	\centering
	\begin{subfigure}{0.33\linewidth}
		\includegraphics[width=1\columnwidth]{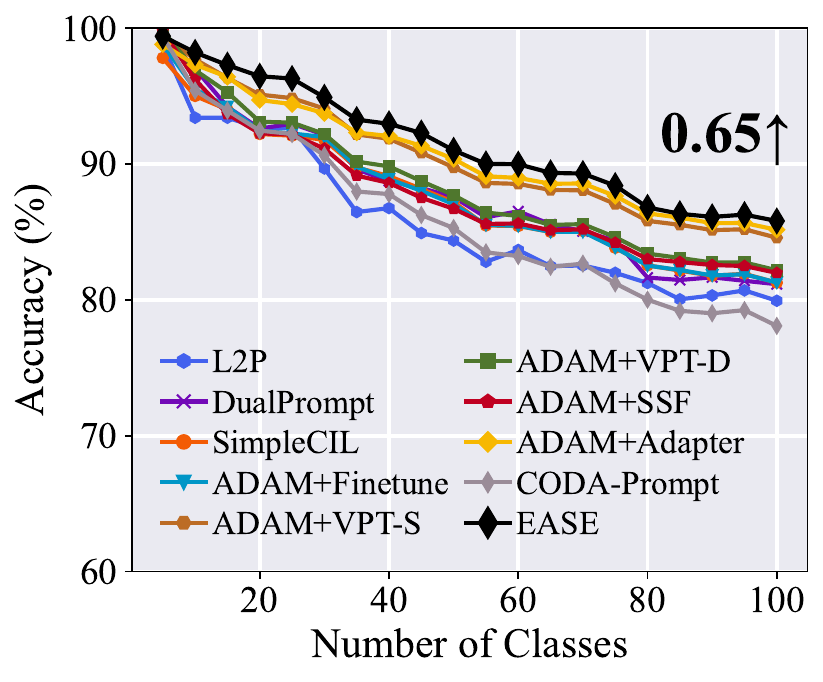}
		\caption{CIFAR B0 Inc5}
	\end{subfigure}
	\hfill
	\begin{subfigure}{0.33\linewidth}
		\includegraphics[width=1\linewidth]{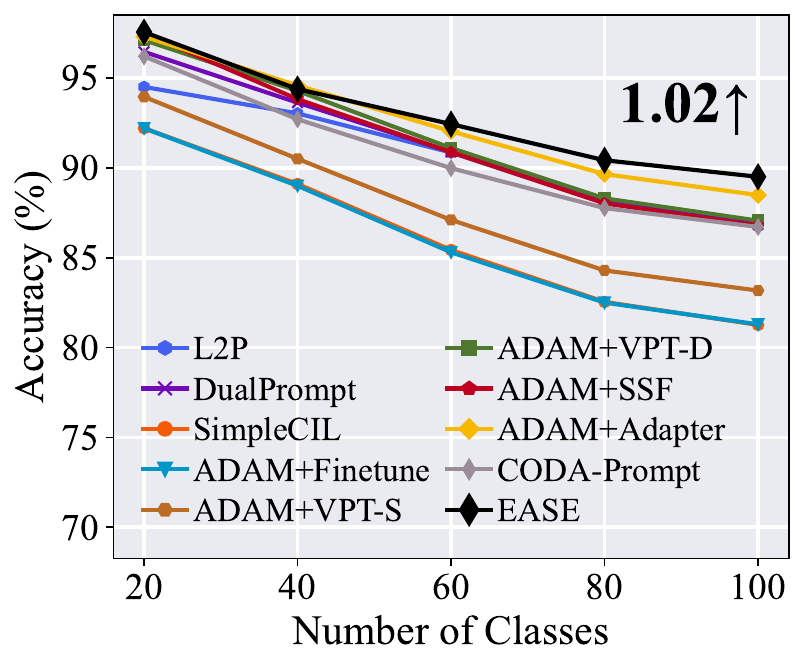}
		\caption{CIFAR B0 Inc20}
	\end{subfigure}
	\hfill
	\begin{subfigure}{0.33\linewidth}
		\includegraphics[width=1\linewidth]{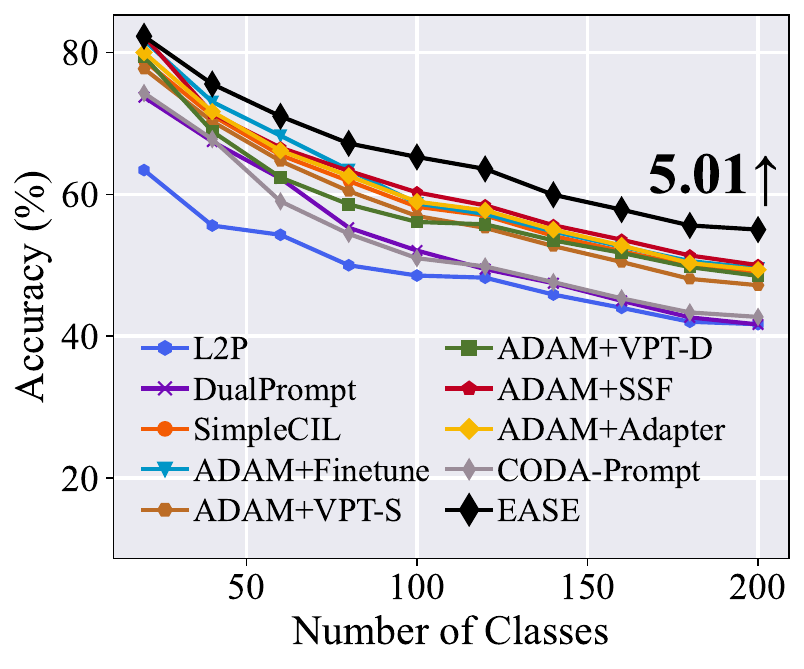}
		\caption{ImageNet-A B0 Inc20}
	\end{subfigure}
	\\
	\begin{subfigure}{0.33\linewidth}
		\includegraphics[width=1\linewidth]{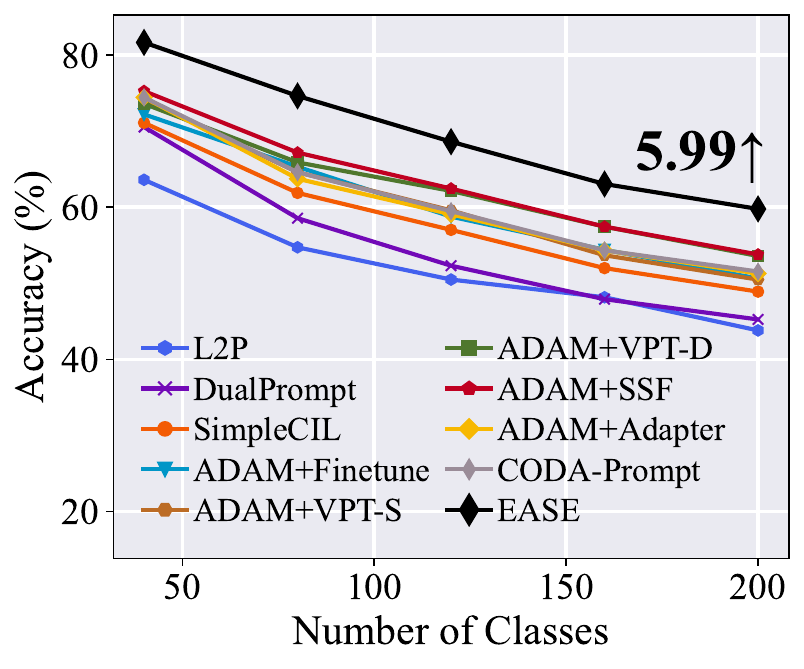}
		\caption{ImageNet-A B0 Inc40}
	\end{subfigure}
	\hfill
	\begin{subfigure}{0.33\linewidth}
		\includegraphics[width=1\linewidth]{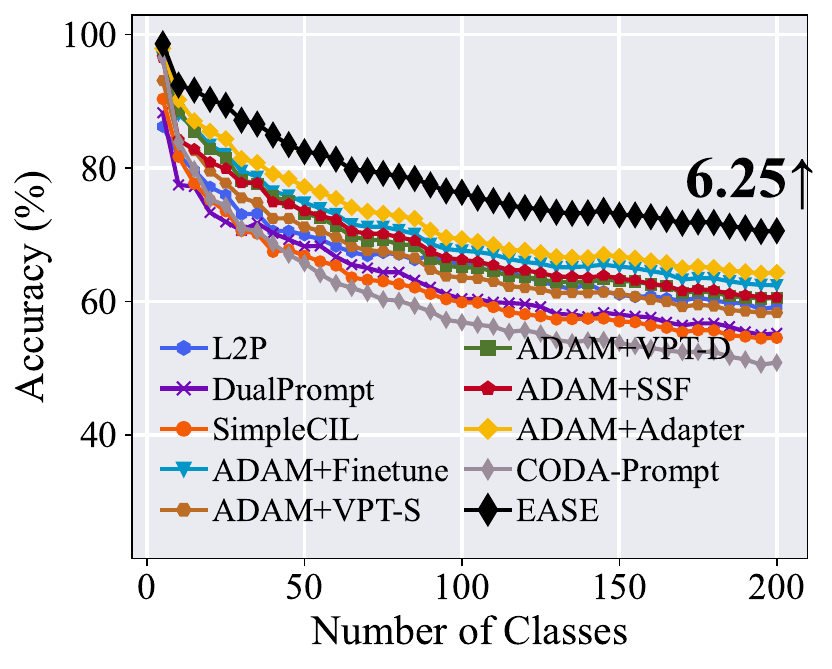}
		\caption{ImageNet-R B0 Inc5}
	\end{subfigure}
	\hfill
	\begin{subfigure}{0.33\linewidth}
		\includegraphics[width=1\columnwidth]{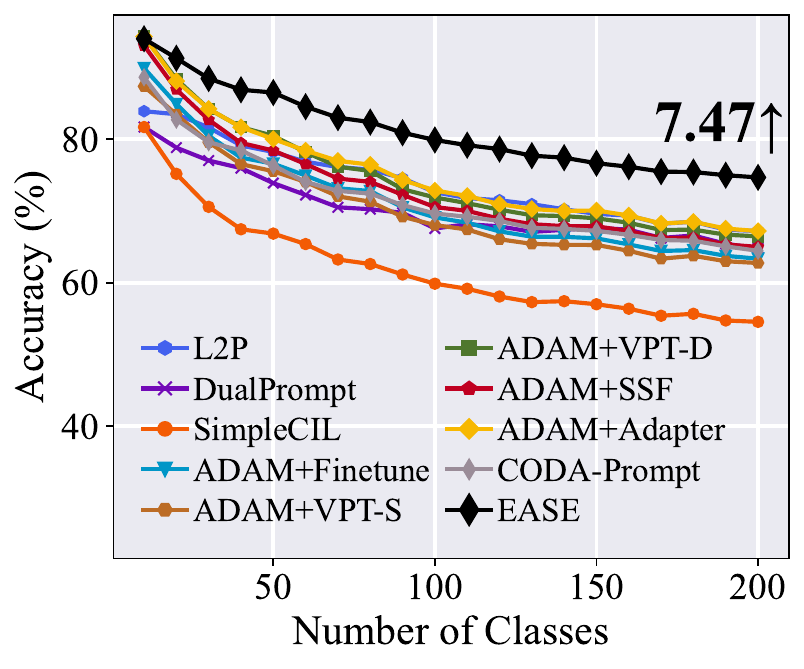}
		\caption{ImageNet-R B0 Inc10}
	\end{subfigure}
	\\
	\begin{subfigure}{0.33\linewidth}
		\includegraphics[width=1\linewidth]{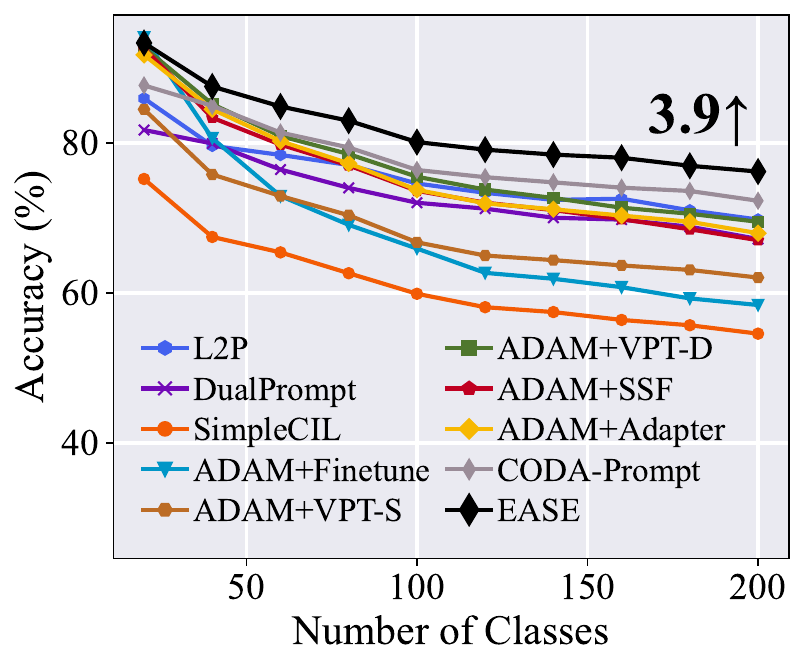}
		\caption{ImageNet-R B0 Inc20}
	\end{subfigure}
	\hfill
	\begin{subfigure}{0.33\linewidth}
		\includegraphics[width=1\linewidth]{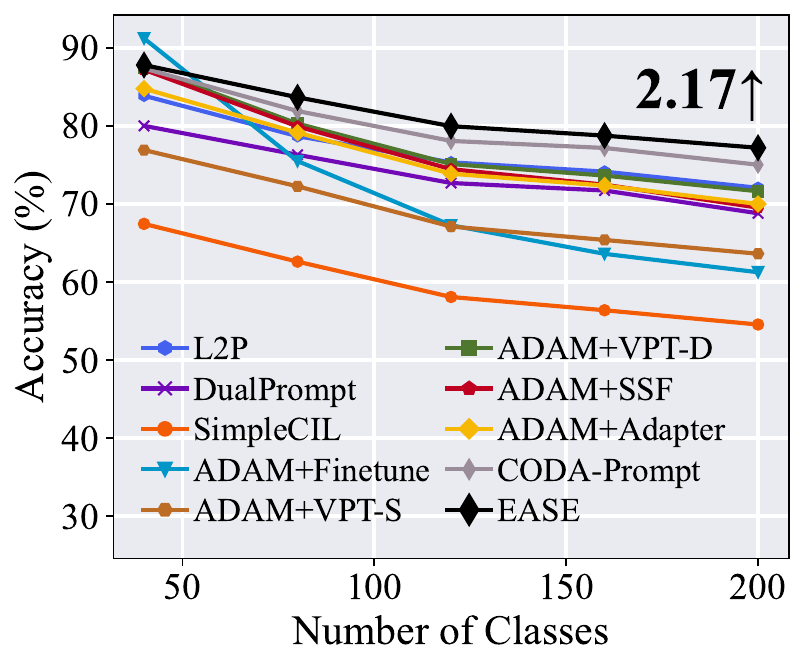}
		\caption{ImageNet-R B0 Inc40}
	\end{subfigure}
	\hfill
	\begin{subfigure}{0.33\linewidth}
		\includegraphics[width=1\linewidth]{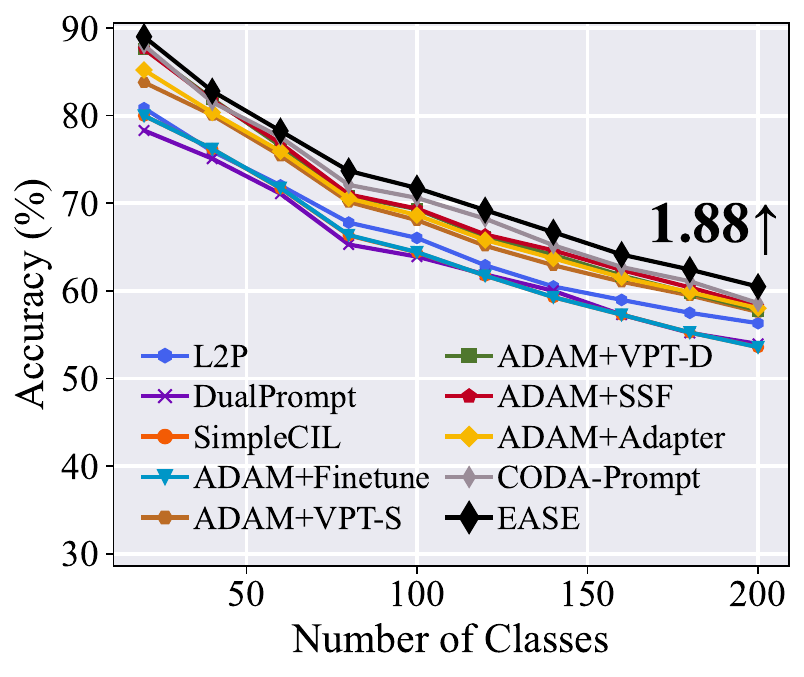}
		\caption{ObjectNet B0 Inc20}
	\end{subfigure}
	\\
	\begin{subfigure}{0.33\linewidth}
		\includegraphics[width=1\linewidth]{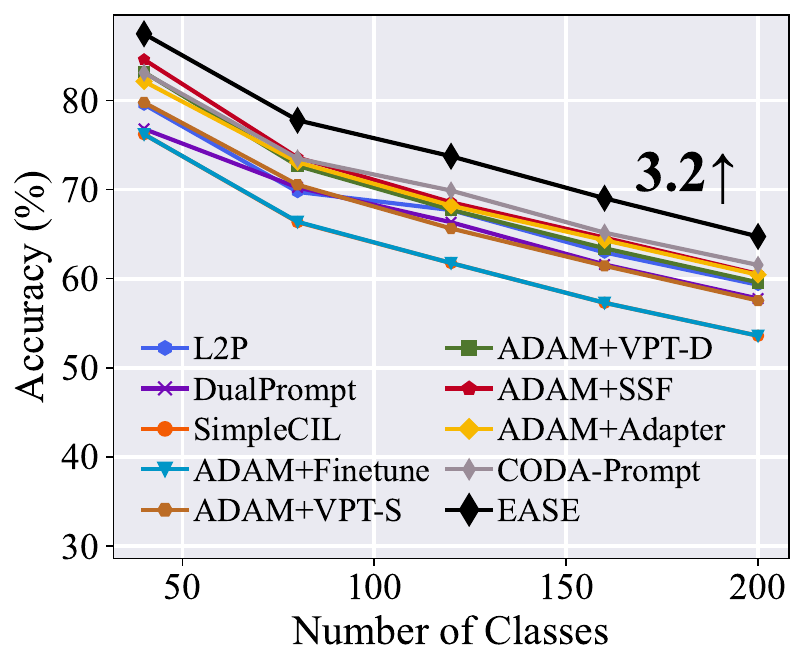}
		\caption{ObjectNet B0 Inc40}
	\end{subfigure}
	\hfill
	\begin{subfigure}{0.33\linewidth}
		\includegraphics[width=1\columnwidth]{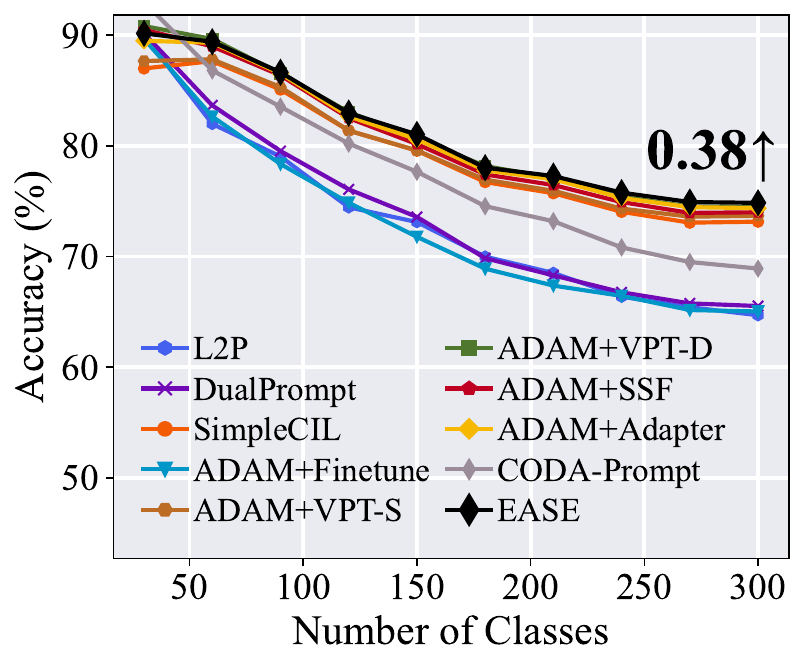}
		\caption{OmniBenchmark B0 Inc30}
	\end{subfigure}
	\hfill
	\begin{subfigure}{0.33\linewidth}
		\includegraphics[width=1\columnwidth]{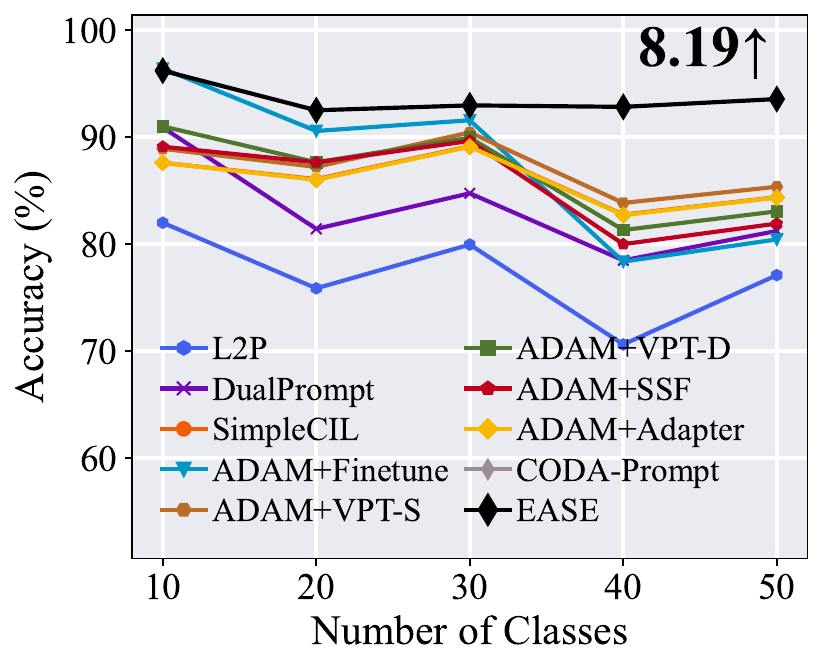}
		\caption{VTAB B0 Inc10}
	\end{subfigure}
	\caption{ Performance curve of different methods under different settings. All methods are initialized with {\bf ViT-B/16-IN21K}. We annotate the relative improvement of \name above the runner-up method with numerical numbers at the last incremental stage. }
	\label{fig:supp-in21k}
\end{figure*}

\section{Introduction About Compared Methods}\label{sec:supp_intro}

In this section, we introduce the details of compared methods adopted in the main paper. {\bf All methods are based on the same pre-trained model for a fair comparison.} They are listed as:

\begin{itemize}
	\item {\bf Finetune}: with a pre-trained model as initialization, it finetunes the PTM with cross-entropy loss for every new task. Hence, it suffers sever catastrophic forgetting on former tasks.
	\item {\bf LwF~\cite{li2017learning}}: aims to utilize knowledge distillation~\cite{hinton2015distilling} to resist forgetting. In each new task, it builds the mapping between the last-stage model and the current model to reflect old knowledge in the current model.
	\item {\bf SDC~\cite{yu2020semantic}}: utilizes a prototype-based classifier. During model updating, the feature drifts, and the old prototypes cannot represent former classes. Hence, it utilizes new class instances to estimate the drift of old classes.
	\item {\bf L2P~\cite{wang2022learning}}: is the first work introducing pre-trained vision-transformers into continual learning. During model updating, it freezes the pre-trained weights and utilizes visual prompt tuning~\cite{jia2022visual} to trace the new task's features. It builds instance-specific prompts with a prompt pool, which is constructed via key-value mapping.
	\item  {\bf DualPrompt~\cite{wang2022dualprompt}}: is an extension of L2P, which extends the prompt into two types, \ie, general and expert prompts. The other details are kept the same with L2P, \ie, using the prompt pool to build instance-specific prompts.
	\item {\bf CODA-Prompt~\cite{smith2023coda}}: noticing the drawback of instance-specific prompt select, it aims to eliminate the prompt selection process by prompt reweighting. The prompt selection process is replaced with an attention-based prompt recombination.
	\item {\bf SimpleCIL~\cite{zhou2023revisiting}}: explores prototype-based classifier with vanilla pre-trained model. With a PTM as initialization, it builds the prototype classifier for each class and utilizes a cosine classifier for classification.
	\item {\bf ADAM~\cite{zhou2023revisiting}}: extends SimpleCIL by aggregating the pre-trained model and adapted model. It treats the first incremental stage as the only adaptation stage and adapts the PTM to extract task-specific features. Hence, the model can unify generalizability and adaptivity in a unified framework.
	
\end{itemize}

{\bf Above methods are exemplar-free, which do not require using exemplars. However, we also compare some exemplar-based methods in the main paper as follows:}

\begin{itemize}
	
	\item {\bf iCaRL~\cite{rebuffi2017icarl}}: utilizes knowledge distillation and exemplar replay to recover former knowledge. It also utilizes the nearest center mean classifier for final classification.
	\item {\bf DER~\cite{yan2021dynamically}}: explores network expansion in class-incremental learning. Facing a new task, it freezes the prior backbone to keep it in memory and initializes a new backbone to extract new features for the new task. With all historical backbones in the memory, it utilizes the concatenation as feature representation and learns a large linear layer as the classifier. The linear layer maps the concatenated features to all seen classes, requiring exemplars for calibration. DER shows impressive results in class-incremental learning, while it requires large memory costs for saving all historical backbones.
	\item {\bf FOSTER~\cite{wang2022foster}}: to alleviate the memory cost of DER, it proposes to compress backbones via knowledge distillation. Hence, only one backbone is kept throughout the learning process, and it achieves feature expansion with low memory cost.
	\item {\bf MEMO~\cite{zhou2022model}}: aims to alleviate the memory cost of DER from another aspect. It decouples the network structure into specialized (deep) and generalized (shallow) layers and extends specialized layers based on the shared generalized layers. Hence, the memory cost for network expansion decreases from a whole backbone to generalized blocks. In the implementation, we follow~\cite{zhou2022model} to decouple the vision transformer at the last transformer block.
	
\end{itemize}

In the experiments, we reimplement the above methods based on their source code and PyCIL~\cite{zhou2021pycil}.

\section{Full Results} \label{sec:supp_full_results}

In this section, we show more experimental results of different methods. 
Specifically, we report the incremental performance of different methods with ViT-B/16-IN21K  in Figure~\ref{fig:supp-in21k}. 
As shown in these results, \name consistently outperforms other methods on different datasets by a substantial margin.

\section{Pseudo Code}

{
	\begin{algorithm}[t]
		\small
		\caption{ \name for CIL }
		\label{alg1}
		\raggedright
		{\bf Input}: Incremental datasets: $\left\{\D^{1}, \D^{2}, \cdots, \D^{B}\right\}$, Pre-trained embedding: $\phi(\x)$;\\
		{\bf Output}: Incrementally trained model; 
		\begin{algorithmic}[1]
			\For{$b=1,2\cdots,B$}  
			\State Get the incremental training set $\D^b$; 
			\State Initialize a new adapter $\A_b$;
			\State Optimize the subspace via Eq.~5; \label{line:optimize-adapter}
			\State Extract the prototypes of $\D^b$ for all adapters via Eq.~7; 
			\State Complete the prototypes for former classes via Eq.~9; \label{line:complete-prototype}
			\State Construct the prototypical classifier via Eq.~10;
			\State Test the model via Eq.~12; \label{line:reweight-logit}
			\EndFor 
			\Return the updated model;
		\end{algorithmic}
	\end{algorithm}
}
We summarize the training pipeline of \name in Algorithm~\ref{alg1}. We initialize and train an adapter for each incoming task to encode the task-specific information (Line~\ref{line:optimize-adapter}). 
Afterward, we extract the prototypes of the current dataset for all adapters and synthesize the prototypes of former classes (Line~\ref{line:complete-prototype}). Finally, we construct the full classifier and reweight the logit for prediction (Line~\ref{line:reweight-logit}).
Since we are using the prototype-based classifier for inference, the classifier $W$ in Eq.~5 will be dropped after each learning stage.